\documentclass[arxiv]{melba}

\usepackage{amsmath,amsfonts}

\usepackage[nolist]{acronym}
\usepackage{svg}
\usepackage{subcaption}
\usepackage{xurl}
\usepackage[figuresright]{rotating}
\usepackage{mathtools}
\DeclarePairedDelimiter\floor{\lfloor}{\rfloor}

\melbaid{YYYY:NNN}  
\doi{10.59275/j.melba.2024-AAAA}
\melbaauthors{Keuth, Kaftan, Heinrich}  
\email{r.keuth@uni-luebeck.de}
\volume{3}
\firstpageno{1337}  
\melbayear{YYYY}  
\datesubmitted{2025-10-20}  
\datepublished{yyyy-m2-d2}  

\ShortHeadings{Shaken or Stirred?}{Keuth, Kaftan, Heinrich}

\title{Shaken or Stirred? An Analysis of MetaFormer's Token Mixing for Medical Imaging}

\author{
	\firstname Ron \surname Keuth\aff{1}\orcid{0009-0003-4289-2836},
    \firstname Paul \surname Kaftan\aff{2}\orcid{0009-0009-0856-586X},
    \firstname Mattias P. \surname Heinrich\aff{1}\orcid{0000-0002-7489-1972}
}
\affiliations{
	\num 1 \addr Institute of Medical Informatics, University of Lübeck, Lübeck, Germany \\
	\num 2 \addr Institute of Medical Systems Biology, Ulm University, Ulm, Germany
}

\abstract{
    The generalization of the Transformer architecture via MetaFormer has reshaped our understanding of its success in computer vision.
    By replacing self-attention with simpler token mixers, MetaFormer provides strong baselines for vision tasks.
    However, while extensively studied on natural image datasets, its use in medical imaging remains scarce, and existing works rarely compare different token mixers, potentially overlooking more suitable designs choices.
    In this work, we present the first comprehensive study of token mixers for medical imaging.
    We systematically analyze pooling-, convolution-, and attention-based token mixers within the MetaFormer architecture on image classification (global prediction task) and semantic segmentation (dense prediction task).
    Our evaluation spans nine datasets (seven 2D and two 3D) covering diverse modalities and common challenges in the medical domain.
    Given the prevalence of pretraining from natural images to mitigate medical data scarcity, we also examine transferring pretrained weights to new token mixers.
    Our results show that, for classification, low-complexity token mixers (e.g. grouped convolution or pooling) are sufficient, aligning with findings on natural images.
    Pretrained weights remain useful in some settings despite the domain gap introduced by the new token mixer.
    For segmentation, we find that the local inductive bias of convolutional token mixers is essential.
    Grouped convolutions emerge as the preferred choice, as they reduce runtime and parameter count compared to standard convolutions, while the MetaFormer's channel-MLPs already provide the necessary cross-channel interactions.
    Our code is available at~\url{https://github.com/multimodallearning/MetaFormerMedImaging/tree/clean_code}.
}


\begin{document}
\begin{acronym}
\acro{cnn}[CNN]{Convolutional Neural Network}
\acro{nlp}[NLP]{Natural Language Processing}
\acro{vit}[ViT]{Vision Transformer}
\acro{mlp}[MLP]{Multi-Layer Perceptron}
\acro{sar}[SAR]{Synthetic Aperture Radar}
\acro{ct}[CT]{Computer Tomography}
\acro{mri}[MRI]{Magnetic Resonance Imaging}
\acro{flops}[FLOPs]{Floating Point Operations}
\acro{auc}[AUC]{Area under the Receiver Operating Characteristic curve}
\acro{dsc}[DSC]{Dice Similarity Score}
\acro{detr}[DETR]{Detection Transformer}
\acro{gap}[GAP]{Global Average Pooling}
\end{acronym}

\twocolumn[\maketitle]

\section{Introduction}
\enluminure{O}{ver} the last ten years, Deep Learning and especially \acp{cnn} have set the state-of-the-art in various tasks of medical image processing like classification \citep{anwar2018medicalCNN}, segmentation \citep{isenseeNNUNet2021} and detection \citep{CNNMedicalDetection}.
Following its success in \ac{nlp}~\citep{devlin2019bert, radford2019gpt}, self-attention \citep{vaswani2023attentionneed} has soon emerged as an alternative to convolution.
The Transformer architecture has been successfully adapted to image classification \citep{dosovitskiy2021vit}, segmentation \citep{xie2021segformer} and detection \citep{carion2020detr}.
Further adaptation of the Transformer architecture targeted specific requirements for vision tasks.
This includes for example a hierarchical pyramid-like architecture like it can be found in \acp{cnn} \citep{wang2021pyramidTransformer, liu2021swintransformer}, the limitation of its receptive field to the direct neighbourhood for detection \citep{zhu2020deformableDETR}, or tackling characteristics of self-attention that hinder the usage of \acp{vit} as the backbone for further vision tasks \citep{darcet2023VITregisters}.\par
The property of self-attention having a global receptive field is a doubled-edged sword in image processing where large input sizes are common.
On the one hand, the number of trainable parameters of the self-attention operator is independent of the input size, and it enables capturing of large context.
On the other hand, since every pixel attends to each other pixel in an input, its \emph{computational complexity} grows quadratically with the resolution of 2D images, making it inefficient for high resolutions.
In contrast, convolution limits its receptive field to the direct neighbourhood of the pixel and thus provides a strong local bias and translation-equivariance, two properties fitting the domain of images.
Combining both paradigms, many works have also limited the receptive field of self-attention, introducing \emph{local} self-attention.
Here, the implementations can be divided into two groups, splitting the input into patches \citep{vaswani2021scalingLocalAttention} or limiting the influence of pixels outside the direct neighbourhood by applying a (soft) weighting during the self-attention operation \citep{d2021convit}.
For both local and global attention, implementations were introduced \citep{pan2023slideAttention, dao2022flashattention}.
Further, recent work investigates into the principle of filter separation, bringing state space models to computer vision within the MAMBA architecture \citep{zhu2024visionmamba}.
Lately, frameworks like PyTorch's \texttt{flexattention} enables the easy implementation of various attention patterns.\par
There also exist efforts (e.g. classification \citep{liu2022convnext} and segmentation \citep{lee2022UXnet, roy2023mednext}) to enable large receptive fields for \acp{cnn} without equipping the model with too many learnable parameters.\par
While \cite{tolstikhin2021mlpmixer} demonstrate that replacing the self-attention within the \ac{vit} with \ac{mlp} yields comparable performance \citep{dosovitskiy2021vit, vaswani2021scalingLocalAttention}, the MetaFormer framework \citep{yu2022metaformeractuallyneedvision} takes this a step further by abstracting the general transformer architecture into two parts: The \emph{token mixer} (e.g. self-attention) learns to aggregate spatial information and the \emph{channel-\acs{mlp}} extracts new features with a point-wise operation.
Within this MetaFormer perspective, using grouped convolution as the token mixer naturally draws a parallel to ConvNeXt \citep{liu2022convnext}, where depthwise convolution fulfils an analogous role in aggregating spatial information.
However, using pooling as a simple, non-parameterised token mixer, the MetaFormer outperforms \ac{vit} on ImageNet.
This implies that the channel-\acp{mlp} provide the main contribution to the transformer's capabilities in computer vision, rather than the token mixer.
In their following work, \cite{yu2023metaformer} provide strong results (ImageNet top-1 accuracy of over $80\,\%$) using no token mixing at all.
Still, equipping the token mixers with some learnable parameters was found beneficial (see \cite{yu2022metaformeractuallyneedvision}, Tab.~5).\par
The general formulation of the MetaFormer easily allows for the deployment of hybrid architectures, combining different token mixers (e.g. using pooling as an efficient token mixer in the first stages with high-resolution images and global self-attention in latter ones).
Since then, MetaFormer has been implemented in various applications and extended with specialised token mixers (see Sec. \ref{sec:related_work}).

\subsection{Contribution}
Most existing applications of the MetaFormer lie outside the medical domain, and those within typically employ it without comparing different token mixers -- potentially overlooking more suitable choices.
The medical image domain has fundamental differences to natural images with relatively smaller available datasets, different semantics of texture vs. shape features, and often standardized viewpoints with possibly subtle feature changes between scanners.
Inductive biases of token mixers will impact results differently in both domains.
This work presents the first comprehensive study of token mixers for medical imaging, analysing their strengths, weaknesses, and complexity in terms of trainable parameters and runtime.
We systematically compare seven token mixers, varying the kernel size for kernel-based mixers such as convolution and pooling.
Our evaluation covers nine datasets, comprising seven 2D and two 3D tasks.
We investigate two different tasks: classification as a global prediction task (five datasets) and semantic segmentation as a dense prediction task (four datasets, covering small, medium, and large resolutions).
The datasets represent diverse modalities, including histology, microscopy, and radiography.
With this, we chose to expose the different token mixers to key medical imaging challenges such as large homogeneous regions in abdominal CT and long-range context requirements in pathology.
We summarise our key findings as follows.
\begin{itemize}
    \item Datasets containing a single dominant object in the image favour more locality and thus smaller kernel sizes. Conversely, datasets with a larger region of interest tend to profit from incorporating a broader spatial context through larger kernels.
    \item For classification, the MetaFormer trained from scratch with no token mixing already yields very competitive performance. The results are best when using low-complexity token mixers.
    \item When pretrained weights are available, exchanging token mixers introduces a domain gap. But pretrained weights of global attention can still be successfully utilized in local attention, provided the data domain remains consistent. This provides a more efficient alternative to global attention when dealing with large input resolutions. When an additional domain shift is introduced -- such as moving to medical imaging -- the benefits become less consistent, and improvements are only observed for certain token mixers and datasets.
    \item For segmentation, we observe that a strong local inductive bias is crucial, making grouped convolutional token mixers the most effective choice.
\end{itemize}

\section{Related Works} \label{sec:related_work}
With the right choice of token mixer, the MetaFormer can be a light-weighted model, enabling its use for real-time applications like ship signature classification in \acf{sar} \citep{zhu2023sipclassification} or in autonomous driving \citep{shun2024efficientMetaFormer}, which deploys additional token pruning for reduced runtime complexity.
For super resolution of meteorological satellite cloud images, \cite{wang2023superresolution} introduce a new token mixer combining gated convolution and channel-space attention.
With the extension of temporal pooling, the MetaFormer has been applied on time series, capturing the structural and temporal relationship in human motion prediction for human-robot interaction \citep{xu2023humaninteraction}.
A novel idea of token mixing moves this operation to the Fourier spectrum, enabling efficient global token mixing without limited receptive field, but adding the cost of the Fourier transformation and its inverse.
With balancing the low- and high-pass characteristics via spectral pooling aggregation \citep{yun2023spanet}, the performance on ImageNet is improved by $3\,\%$ accuracy compared to the PoolFormer (MetaFormer with average pooling as token mixer) \citep{yu2022metaformeractuallyneedvision}.
In \cite{tatsunami2024fftMetaformer}, this approach is extended, so coefficients of a dynamic filter bank are estimated by an \ac{mlp} based on the input.
While the original work \citep{yu2022metaformeractuallyneedvision} employs the MetaFormer itself as a backbone for semantic segmentation, MetaSeg \citep{Kang2024MetaSeg} extends its principles to a decoder and introduces channel reduction attention for gathering global context within the decoder.\par

In the medical domain, MetaFormer mostly gets adapted for semantic segmentation.
For polyp segmentation, MetaFormer has been paired with a parallel path comprising convolutional units with residual and dilated paths to compensate for MetaFormer's weak low-level features \citep{MetaFormerCNNPolypSegmentation}.
Another work on polyp segmentation \citep{trinh2023m2unet} utilizes the MetaFormer as the encoder in a U-Net-like architecture and extend the decoder's dilated convolutions with learnable multiscale upsampling.
In volumetric brain tumour segmentation, MetaFormer has been used as an auxiliary decoder, combining local–global pooling with depthwise convolution as the token mixer.
To improve efficiency, it has been adapted with involution and a squeeze-and-excitation module in place of channel-\acs{mlp} \citep{GhostNetMobileLightweightedSegmentation}.
Similarly, MAMBA’s structured visual space model has been integrated into MetaFormer for image restoration of brain \ac{mri} and abdominal \ac{ct} \citep{MetaMambaImageRestouration}.
For pancreas tissue detection, deformable convolutions have been employed as token mixers, dynamically adapting the receptive field \citep{DeformableFormer}.
To the best of our knowledge, this is the first comprehensive study of different standard token mixers in the MetaFormer on medical image analysis.

\section{Methods}

\begin{figure}
    \centering
    \includegraphics[height=.3\textheight]{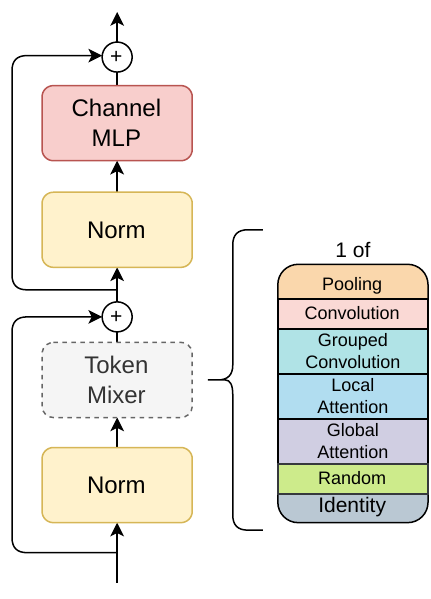}
    \caption{Schema of the MetaFormer block illustrating different token mixer options employed across stages of the MetaFormer architecture (cf. Fig.~\ref{fig:architetcure}).}
    \label{fig:metaformerblock}
\end{figure}
\begin{figure*}
    \centering
    \begin{subfigure}{\linewidth}
        \centering
        \includegraphics[width=.85\linewidth]{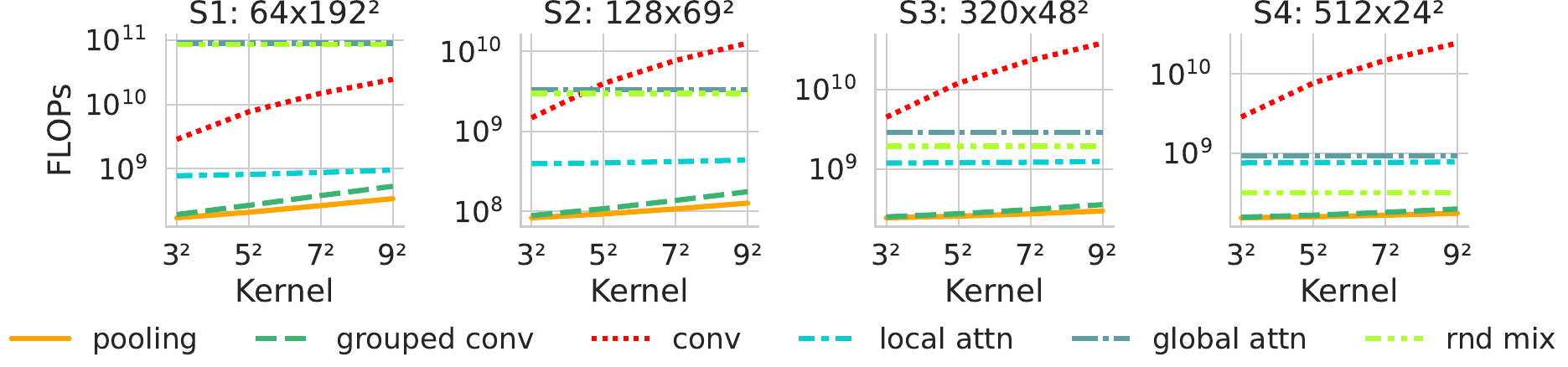}
        \caption{\acs{flops} of the token mixers across the four MetaFormer stages (stage index: $C \times N$) for an input of size $3 \times 768^2$.}
        \label{fig:flops}
    \end{subfigure}
    \begin{subfigure}{\linewidth}
        \centering
        \includesvg[width=.8\linewidth]{img/Architecture.svg}
        \caption{Extension of MetaFormer for classification by collapsing the final feature map to a vector and segmentation employing the SegFormer decoder \citep{xie2021segformer} with number of classes $|\Omega|$ and latent dimension $C=256$ for segmentation.}
        \label{fig:architetcure}
    \end{subfigure}
    \caption{Visualization of \acs{flops}  for different token mixers (\subref{fig:flops}) and schema of MetaFormer's architecture (\subref{fig:architetcure}).}
\end{figure*}

In a typical medical image processing task, we are given an input image $\mathbf{I}\in\mathbb{R}^{C\times H\times W}$ and aim to learn a function $f(\mathbf{I})$ that projects the image into a predefined solution space.
In image classification, a vector in this solution space is described as a global vector $\mathbf{s}\in\mathbb{R}^{|\Omega|}$ holding the likelihood $p(\mathbf{I}|\omega_n)$ for each class $\omega\in\Omega$.
For semantic segmentation, the classification vector is defined for every pixel $\mathbf{p}\in\mathbf{I}$.
Thus, we obtain a segmentation mask $\mathbf{S}\in\mathbb{R}^{|\Omega|\times H\times W}$ holding the likelihoods for every pixel $p(\mathbf{p}|\omega)$ to belong to class $\omega$.
When approximating the optimal $f$ in a given task with deep learning, we used to implement $f=f_0\circ f_1 \circ \dots \circ f_L$ as a nested chain of functions $f_l$, where each $\{f_l\}_{l=0}^L$ represent a layer in a neural network.
In the past ten years, the favoured operator in image processing for $f_l$ has been convolution \citep{AlexNet, he2015resnet, liu2022convnext}.
After their initial success in natural language processing tasks, Transformers with their self-attention operator have also emerged as an alternative to convolutional networks \citep{dosovitskiy2021vit, xie2021segformer, carion2020detr}.
However, while \acp{vit} achieve state-of-the-art performance in many image tasks, it remains unclear, which architectural components drive their success. 
To address this, MetaFormer \citep{yu2022metaformeractuallyneedvision, yu2023metaformer} generalizes the Transformer architecture by replacing self-attention with operations better suited to vision.

\subsection{MetaFormer Architecture}
The MetaFormer \citep{yu2022metaformeractuallyneedvision} block represents a generalized version of the Transformer in that it defines token mixing as a variable operation, of which self-attention is only one of many possible operations.
The other structures of the Transformers (e.g. channel MLP or residual connections) have been proven to be also crucial for computer vision \citep{yu2022metaformeractuallyneedvision} and are therefore left untouched.
Fig.~\ref{fig:metaformerblock} shows the structure of the MetaFormer's block.
Within its multi-stage architecture, each employing multiple MetaFormer blocks, the MetaFormer adopts the hierarchical approach of \acp{cnn} \citep{he2015resnet, AlexNet} of halving the spatial resolution and doubling the feature channels at the beginning of each stage using a convolution layer (see Fig. \ref{fig:architetcure}).
The first stage is an exception with a downsampling factor of four.
The number of blocks per stage is a hyperparameter to modify model capacity, similar to the ResNet \citep{he2015resnet}.
With this, the MetaFormer enables feature extraction on different resolution scales, which stands in contrast to the original \ac{vit} \citep{dosovitskiy2021vit}, where the image is patchified and then fed as a sequence of patches through the network without deviating from the original resolution.

\subsection{Different Token Mixers}\label{sec:token_mixer}
Following MetaFormer \citep{yu2022metaformeractuallyneedvision, yu2023metaformer}, we investigate commonly used operations as token mixers denoted by $t(x)$.
We compare identity, average pooling, convolution, grouped convolution, self-attention, and local self-attention, all defined in the following paragraphs.
The choice of token mixer introduces fundamental differences in properties such as the number of trainable parameters and runtime complexity.
Moreover, each token mixer carries distinct inductive biases regarding locality, making them more appropriate for particular imaging domains (e.g., those requiring strong local interactions, long-range context, or texture-based representations).\par
The computational complexity and trainable parameter count of the investigated token mixers including the channel-MLP of the MetaFormer block are summarized in Tab.~\ref{tab:token_mixer_complexity}.
We state the complexity in terms of the input size: number of channels $C$, height $H$, and width $W$ (for shorter notation we define $N=HW$).
For local operators, we define their kernel or pooling size $K$, where we only consider kernels of size $(K\times K)$ containing $K^2$ elements, since it is the most common choice.
Note that a MetaFormer block does not change the spatial resolution or number of channels and, therefore, we use the same variables for in- and output sizes.\par
To define the following operations, we introduce the indices $c,h,w\in\mathbb{N}$ iterating over the $C$ channels, the height $H$ and width $W$.
We define the pixel value at the index $(c,h,w)$ as $\mathbf{I}(c, h, w) = p\in\mathbb{R}$.

\subsubsection{Identity}
Although strictly speaking, the identity provides no token mixing, we include it as a lower baseline to investigate which benefits token mixing provides on its own.
\begin{equation}
    t_\texttt{id}\left(\mathbf{I},c,h,w\right)=\mathbf{I}(c,h,w)
\end{equation}
The identity operation introduces neither additional computational cost nor additional parameters to the MetaFormer architecture.

\subsubsection{Average Pooling}
As a next step, we introduce spatial average pooling.
This operation performs basic token mixing and adds a local inductive bias beneficial for image processing, while requiring no additional parameters.
The average is computed within a pool of size $K^2$:
\begin{equation}
    t_\texttt{avg}\left(\mathbf{I},c,h,w\right)=\frac{1}{K^2}\sum_{i=-\floor{\frac{K}{2}}}^{\floor{\frac{K}{2}}}\sum_{j=-\floor{\frac{K}{2}}}^{\floor{\frac{K}{2}}}\mathbf{I}(c,h+i,w+j)
\end{equation}
With this, its computational complexity is $\mathcal{O}(NK^2C)$.

\subsubsection{Convolution}
Convolution provides with its locality a favourable inductive bias for data on regular grids like images with highly correlated features in neighbourhoods.
By optimizing its kernel parameters $\mathbf{K}\in\mathbb{R}^{C\times C\times K\times K}$ by gradient descent \citep{LeCun1989}, it has been responsible for the breakthrough of Deep Learning in computer vision, providing a higher efficiency through its sparsity and reuse of parameter when compared to previously standard fully connected models.
All this makes convolution a natural choice as a token mixer in image-based tasks.
\begin{align}\label{eq:conv}
    t_\texttt{conv}&\left(\mathbf{I},\mathbf{K},c_{\texttt{out}},h,w\right)\\
    &=\sum_{c=1}^C\sum_{i=-\floor{\frac{K}{2}}}^{\floor{\frac{K}{2}}}\sum_{j=-\floor{\frac{K}{2}}}^{\floor{\frac{K}{2}}}\mathbf{I}(c,i,j)\mathbf{K}(c_{\texttt{out}},c,h+i,w+j)\notag
\end{align}
Due to its connectivity over the channels, convolution has a high computational complexity of $\mathcal{O}(NK^2C^2)$ and uses $K^2C^2$ parameters, mostly limiting the choice of kernel size to $K=3$ for all but the first layers \citep{AlexNet, he2015resnet}. 

\subsubsection{Grouped Convolution}
To overcome the high complexity of convolution, grouped convolution can be used \citep{AlexNet}.
Using $G$ filter groups, an output channel no longer depends on all input channels but on a subset of $\frac{C}{G}$.
Thus, the depth connectivity can be reduced ($G<C$) or omitted ($G=C$).
Note that $G$ has to be chosen so that $C$ is an integer multiple of $G$.
To show the extrema, we chose $G=C$, shrinking the kernel weight matrix to $\mathbf{K}\in\mathbb{R}^{C\times K\times K}$ and thus omit the sum over the channel (compare Eq.~\ref{eq:conv})
\begin{align}
    t_\texttt{group\_conv}&\left(\mathbf{I},\mathbf{K},c,h,w\right)\\
    &=\sum_{i=-\floor{\frac{K}{2}}}^{\floor{\frac{K}{2}}}\sum_{j=-\floor{\frac{K}{2}}}^{\floor{\frac{K}{2}}}\mathbf{I}(c,i,j)\mathbf{K}(c,h+i,w+j)\notag
\end{align}
This reduces complexity to $\mathcal{O}(NK^2C)$ and $K^2C$ parameters.
Since channels are processed independently, grouped convolution with $G=C$ lacks channel cross-talk.
However, as each MetaFormer block includes a channel-MLP (Fig. \ref{fig:metaformerblock}) introducing cross-talk, we argue this limitation is negligible.
This design parallels the depthwise separable convolution of MobileNet \citep{Howard2017_MobileNet}.

\subsubsection{Self-Attention}\label{sec:self-attention}
With Self-Attention, Transformers have set a new state-of-the-art in natural language processing \citep{vaswani2023attentionneed} and have also been well adapted to image processing tasks \citep{dosovitskiy2021vit, xie2021segformer, carion2020detr}.
However, \cite{yu2022metaformeractuallyneedvision} provide evidence that the self-attention mechanism may not be the most important part of a successful model architecture for computer vision.
To test this observation in the medical imaging domain, we include self-attention as a baseline for token mixing.\par
Let $\mathbf{p}_r=\mathbf{I}(h,w) = \mathbf{I}(\lfloor \tfrac{r}{W}\rfloor,r\mod W)\in\mathbb{R}^C$ define the current pixel's feature vector with $r=h\cdot W+w\leq N$ as the ravelled, flat index of $(h,w)$.
For self-attention as token mixer $t_\texttt{global\_attn}$, each pixel $\mathbf{p}_r\in\mathbb{R}^C$ is transformed into the three roles: key $\mathbf{k}_r=\mathbf{W}_k\mathbf{p}_r$, value $\mathbf{v}_r=\mathbf{W}_v\mathbf{p}_r$, and query $\mathbf{q}_r=\mathbf{W}_q \mathbf{p}_r$, where each $\mathbf{W}\in\mathbb{R}^{C\times C}$ describes a learnable weight matrix.
The transformed value of $\mathbf{p}_r$ is obtained by
\begin{align}
    \mathbf{z}_{m,r}&=\sum_{i=1}^N\underset{j=1,\dots,N}{\texttt{softmax}}\left(\frac{\mathbf{q}_{m,r}^\intercal\mathbf{k}_{m,j}}{\sqrt{D}}\right)\cdot\mathbf{v}_{m,i} \label{eq:mha} \\ 
    \mathbf{z}_r&=\mathbf{W}_u\texttt{concat}(\{z_{m,r}\}_{m=1}^M),
\end{align}
where $\mathbf{k},\mathbf{v}$, and $\mathbf{q}$ is split up into $M$ vectors of size $D$, forming attention heads $m=1,\dots,M$, allowing to learn different attention patterns.
Their results are finally unified to $\mathbf{z}_r$ by $\mathbf{W}_u\in\mathbb{R}^{C\times C}$.
In our experiments, we adapt the number of heads $M=\frac{C}{16}$ dynamically for each stage, dividing the number of channels $C$ by 16.
Since self-attention is permutation-invariant, but images have an informative locality structure, we add positional embedding following \cite{yu2022metaformeractuallyneedvision}.
While self-attention's parameter count $4C^2$ is independent of the amount of processed pixels, its computational complexity quadratically grows with their amount $\mathcal{O}(N^2C)$.
This significantly limits the applicability of self-attention for large input resolutions.

\subsubsection{Local Self-Attention}
To break up the quadratic runtime complexity and to infuse self-attention with the same advantages of convolution's locality (e.g. inductive bias and translation equivariance), we limit the number of keys a query can attend to its direct neighbourhood.
Let $\mathcal{N}$ define a function that returns the neighbouring pixel coordinates of a given location:
\begin{equation}\label{eq:kernel}
\mathcal{N}_K(h,w) = \left\{ 
  \begin{aligned}
    i,j \in \mathbb{N}_0 :\; 
    &(|i - h| < \tfrac{K}{2}) \\
    &\land (|j - w| < \tfrac{K}{2})
  \end{aligned}
\right\},
\end{equation}
where $K$ describes the kernel size in height and width, since we limit ourselves to quadratic kernels.
We equivalently denote the neighbourhood function as $\mathcal{N}_K(r)$ with the flat index $r$.
With it, we extend Eq. (\ref{eq:mha}) to construct $t_{\texttt{local\_attn}}$ as follows:
\begin{equation}\label{eq:local_mha}
    \mathbf{z}_{m,r}=\sum_{i\in\mathcal{N}_K(r)}\underset{j\in\mathcal{N}_K(r)}{\texttt{softmax}}\left(\frac{\mathbf{q}_{m,r}^\intercal\mathbf{k}_{m,j}}{\sqrt{D}}\right)\cdot\mathbf{v}_{m,i}.
\end{equation}
We implement $\mathcal{N}_K(r)$'s attention pattern using PyTorch's \texttt{flexattention} framework\footnote{\url{pytorch.org/blog/flexattention}}.
By restricting the receptive field of $\mathbf{q}$, local self-attention reduces complexity to $\mathcal{O}(NK^2C)$ for typical resolutions ($K^2 \ll N=HW$) compared to global self-attention (cf. Fig.~\ref{fig:flops}), while keeping the parameter count unchanged at $4C^2$.

\subsubsection{Random Mixing}
Random mixing \citep{yu2023metaformer} is a non-learnable global token mixing strategy, where attention weights $\mathbf{A}\in\mathbb{R}^{N\times N}$ are randomly initialized but kept fixed.
Intuitively, this replaces the attention scores computed from key and query representations in self-attention (cf. Sec.~\ref{sec:self-attention}).
Analogously to equation (\ref{eq:mha}), the mixing is performed as follows:
\begin{equation}
    t_\texttt{rnd}(\mathbf{I}, r)=\mathbf{z}_{r}=\sum_{i=1}^N\underset{j=1,\dots,N}{\texttt{softmax}}\left(\mathbf{A}(r, j)\right)\cdot\mathbf{p}_{i}.\\ \label{eq:rnd}
\end{equation}
Note because the attention scores are fixed, we can compute the softmax normalization once during the initialization of $\mathbf{A}$.
Like self-attention, random mixing has a quadratic computational complexity depending on the number of processed tokens $\mathcal{O}(N^2C)$.
While random mixing adds no learnable parameters, storing the fixed attention scores increases the model size (adding $21\,\mathrm{M}$ frozen parameters for MetaFormer S12 variant at an input resolution of $224\times224$).

\begin{table*}
    \caption{2D runtime complexity in $\mathcal{O}$ notation, \acf{flops}, and trainable parameter count of different token mixers in the MetaFormer block (plus channel-MLP) regarding the kernel size $K$, number of in- and output channels $C$, and spatial input size $N=H\cdot W$.
    Note that in MetaFormer, no token mixer changes the spatial resolution or number of feature channels.
    Similarly, we only consider kernels of shape $(K\times K)$ comprising $K^2$ elements.
    \ac{flops} formulas are constructed ignoring technical optimisations in hard- and software.}
    \centering
    \begin{tabular}{l|r|r|r}
        \toprule
       Token Mixer              & Runtime Complexity                & \acs{flops} formula   & \# Parameters  \\
       \midrule
       Identity                 & $\mathcal{O}(NC^2)$               & $NC^2$                & $C^2$\\\addlinespace
       Average Pooling          & $\mathcal{O}(N K^2 C + NC^2)$     & $NK^2C+NC^2$          & $C^2$\\
       Grouped Convolution      & $\mathcal{O}(N K^2 C + NC^2)$     & $N2K^2C + NC^2$       & $K^2C + C^2$ \\
       Local Self Attention     & $\mathcal{O}(NK^2C + NC^2)$       & $5NC^2+NK^2C+N+2NK^2$ & $5C^2$\\\addlinespace
       Convolution              & $\mathcal{O}(N K^2 C^2 + NC^2)$   & $N2K^2C^2+NC^2$       & $K^2C^2 + C^2$\\\addlinespace
       Global Self Attention    & $\mathcal{O}(N^2C + NC^2)$        & $5NC^2+N^2C+N+2N^2$   & $5C^2$\\
       Random Mixing            & $\mathcal{O}(N^2C + NC^2)$        & $N^2C+NC^2$           & $C^2$\\
       \bottomrule
    \end{tabular}
    \label{tab:token_mixer_complexity}
\end{table*}

\section{Experiments}
This section covers our experimental setup, describing the datasets and training routines.
Since medical image datasets are often smaller than natural image datasets like ImageNet, we employ the smallest MetaFormer variant S12 with about $11.4\,\mathrm{M}$ trainable parameters (excluding the token mixers) to carry out our experiments.
It comes with 64, 128, 320 and 512 channels for its four different stages.
Each stage comprises two MetaFormer blocks, except for the third stage (six blocks, see Fig.~\ref{fig:architetcure}).
For both settings, classification and segmentation, we compare the different token mixers introduced in Sec. \ref{sec:token_mixer} utilizing different pooling or kernel sizes where applicable (i.e. not for global attention and identity).
See appendix Sec.~\ref{sec_ap:s_variants}, for an investigation into model scaling using the larger S24 and S36 MetaFormer architecture variants.
Our code is available on \href{https://github.com/multimodallearning/MetaFormerMedImaging/tree/clean_code}{GitHub}\footnote{\url{github.com/multimodallearning/MetaFormerMedImaging/tree/clean_code}}.

\subsection{Classification} \label{sec:class_and_train_routine}
For classification, we extend the MetaFormer encoder with a classification head, averaging the final feature map to a vector via \acf{gap} and then linearly mapping it to the number of classes (see Fig.~\ref{fig:architetcure}).
For optimization, we employ AdamW \citep{Loshchilov2019_Decoupled} with a weight decay of $0.01$ using the default learning rate of $1e^{-3}$, which is reduced by a single cycle of cosine annealing \citep{Loshchilov2017_SGDR} (\texttt{timm} implementation) to a minimum of $1e^{-5}$.
We found it beneficial to employ a linear learning rate warm-up during the first 5 epochs of training.
We adapt the number of epochs to dataset size, targeting 35k iterations with batch size 128.
The final model is selected based on the highest validation F1-score measured at the end of each training epoch.
For the loss function, we chose cross-entropy with label smoothing of 0.1 and class weighting by the inverse root class frequency (clamped to a maximum of 10).
Data augmentation consists of random affine transformations with a transformation matrix sampled from a zero-mean normal distribution with 0.1 standard deviation. 
For additional regularisation, we follow \citep{yu2022metaformeractuallyneedvision} utilizing stochastic depth \citep{Huang2016_Deep} and LayerScale \citep{Touvron2021_Going}.
All images were bilinearly reshaped to $224\times224$.\par

\subsubsection{Pretrained Architecture Variants}
Given the scarcity of medical data, pretrained weights are often essential.
This applies especially for self-attention, which converges slower and requires more data due to its weaker inductive bias compared to convolution \citep{lu2022ViTonSmallData}.
Consequently, we evaluate two MetaFormer variants:
The \emph{first variant} is trained from scratch, with the token mixer of all stages change to the specific experiment setting (denoted as $[T,T,T,T]$).
Since there are no pretrained weights for all our deployed token mixers, we adapted the hybrid architecture of pooling, pooling, global self-attention, global self-attention, for which ImageNet weights have been made public \citep{yu2022metaformeractuallyneedvision}, as our \emph{second pretrained variant}.
Here we keep the pooling in the first two stages and exchange the self-attention in the last two stages according to the experiment setting (denoted as $[P,P,T,T]$) and fine-tune the channel-\acs{mlp} of all stages.
Notably, the third stage contains six layers (vs. two in the others). 
As a consequence, changing the token mixing in this stage modifies half of the token mixer layers of the whole model.
To measure the improvement provided by the pretrained weights, we also train the second variant ($[P,P,T,T]$) from scratch (result shown in appendix Tab.~\ref{tab_ap:2p2t_scratch}).

\subsubsection{Datasets}
We include various classification datasets with vastly different numbers of training images from multiple medical image modalities using the MedMNIST V2 collection \citep{medmnist}.
We select the highest provided image resolution of the datasets at $224\times224$.
From the collection, we select tasks that are harder to classify and thus suitable for comparing token mixers.
We gauge the difficulty of the twelve MedMNIST2D datasets by training a ResNet18 with the aforementioned routine (Sec. \ref{sec:class_and_train_routine}) for up to 500 epochs or $35\,k$ iterations, without a learning rate scheduler.
We then select datasets with a validation F1-score below 0.95 for our studies.
This procedure yielded four diverse datasets:
\emph{PathMNIST} \citep{pathmnist} provides patches from hematoxylin and eosin stained colorectal cancer histological images for the task of classifying nine different types of tissues.
The images are rotation invariant and are split into $89\,996/10\,004/7\,180$ train/validation/test images, making it the largest dataset in our work.
\emph{DermaMNIST} \citep{dermamnist1, dermamnist2} consists of dermatologic images of nine common pigmented skin lesions, leaving it also rotation invariant.
The images are divided into $7\,007/1\,003/2\,005$ splits.
With \emph{PneumoniaMNIST} \citep{Kermany2018_pneumoniamnist}, we include pneumonia detection in chest X-rays with $4\,708 / 524 / 624$ images, respectively.
Lastly, in \emph{OrganSMNIST} \citep{organmnist1, organmnist2} we classify eleven organs in the sagittal view of 201 CT scans, obtaining $13\,932 / 2\,452 / 8\,827$ images respectively.
See Fig.~\ref{fig_ap:medmnist_montages} in appendix for example images of each dataset.\par
As a 3D extension of our investigation, we select NoduleMNIST3D with $1\,158 / 165 / 310$ lung \ac{ct} images for nodule classification \citep{ArmatoIII2011_NoduleMNIST3D}.
We chose this over other MedMNIST 3D datasets, which show either close to saturated performance in benchmarks given in \citep{medmnist} or consist of voxelized shape masks without medical image features.

\subsubsection{CNN Baseline}
As a CNN baseline, we employ ResNet18 \citep{he2015resnet} (\texttt{torchvision} implementation), which matches the parameter count of MetaFormerS12.
For models trained from scratch, we test multiple kernel sizes as in MetaFormer, while for pretrained models we use only kernel size 3, the only available pretrained variant.

\subsubsection{Foundation Model Baseline}\label{sec:foundation_model}
Considering the recent trend of the medical image analysis community towards foundation models \citep{liang2025vfm,ma2024medsam,chen2024uni,perez2025raddino}, we incorporate RAD-DINO \citep{perez2025raddino} as a baseline, leveraging its generalized and discriminative representations from diverse medical imaging datasets.
Following \cite{perez2025raddino}, we train a \ac{mlp} classifier (two layers, 2048 and 256 for hidden and bottleneck dimension, respectively) on the \texttt{CLS} token while keeping the \ac{vit} frozen.
To match the expected input of the \ac{vit}, images are padded to a square aspect ratio and bilinearly upsampled to $518\times 518$.

\subsection{Semantic Segmentation} \label{sec:seg_architecture}
For semantic segmentation, we extract the last block's feature map of all four stages and employ the same light weighted all-\acs{mlp} decoder used in the SegFormer \citep{xie2021segformer} adding $525\,\mathrm{k}$ parameters to the MetaFormer.
We chose this small decoder to minimise its impact on the segmentation tasks and have the MetaFormer be the main feature extractor, amplifying the potential performance difference between types of token mixers.
Fig.~\ref{fig:architetcure} shows a schematic depiction of the decoder.
The SegFormer decoder takes the feature maps from the four MetaFormer stages, equalizes the number of channels to 256 using linear layers, and upsamples them bilinearly to the first stage spatial resolution.
A final two-layer channel-\ac{mlp} with 256 hidden neurons then projects the concatenated $4\cdot256$-channel feature map to the number of classes.
Since, the feature map from the first stage is downsampled by factor four by the MetaFormer's first patch embedding, the logits of the final predicted segmentation mask are bilinearly upsampled by factor four.
It is worth noting, the \acs{mlp}-decoder only includes pointwise operations and thus no token mixing takes place except within the bilinear upsampling (analogous to distance-weighted average pooling).
We use the same training routine as for classification (see Sec. \ref{sec:class_and_train_routine}), only swapping the loss function to cross-entropy plus Dice loss (equally weighted) and setting a fixed number of epochs of 1000 regardless of the dataset size.
Whenever a multiclass segmentation task includes the background class, we ignore it during the loss calculation.

\subsubsection{Datasets}
We use three different medical datasets for semantic segmentation.
The Japanese Society of Radiological Technology (JSRT) Dataset \citep{JSRT} is publicly available and consists of 247 chest X-rays. 
Each image comes with landmark annotations \citep{JSRTAnnotations} for the four anatomical structures of the lungs, heart, and left / right clavicles.
For our experiments, we converted the landmarks to dense masks, downsample the dataset to $256\times256$ pixels, and divide it into a custom split of 160/87 training/test images.\par
As a second dataset, we include bone segmentation of the paediatric wrist based on the publicly available GRAZPEDWRI-DX (GRAZ) dataset \citep{nagyPediatric2022}.
Here we used human expert segmentation mask of 17 bones provided in \cite{keuth2025denseseg} of 63 images, which we downsample to $384\times224$ and split into 43/20 for training/testing.
As the last dataset, we include the Tumor InfiltratinG lymphocytes in breast cancER (TIGER) dataset\footnote{TIGER training data was accessed on July 25th, 2025 from \url{registry.opendata.aws/tiger}.}, comprising large histology slices.
Here, the goal is to segment seven different tissues, including different states/types of tumour and tumour-associated stroma.
We selected the subset of slices where the region of interest is provided with dense segmentation mask, obtaining 284 images split into 256/28 training/testing.
Since the slices have a very high resolution, we follow a patch-based approach with a patch size of $768\times768$.
During training, we sample one patch per image on-the-fly, with probabilities weighted by the inverse square root of class frequency to oversample minority classes.
For inference, we apply a sliding window with $25\,\%$ overlap and merge predictions using Gaussian weighting, reducing the influence of uncertain patch edges.\par
As a 3D segmentation task, we choose AbdomenAtlas 1.0 \citep{li_abdomenatlas_2024} (Abdomen).
The dataset contains CT scans with nine segmented abdominal structures of varying sizes (aorta, gallbladder, left/right kidney, liver, pancreas, postcava, spleen, and stomach).
We employ the official 4675/520 train/test split.
For preprocessing, volumes are cropped to the abdominal region using ground-truth labels, resampled to isotropic $1.5\,\text{mm}$ spacing, padded to resolution of $128^3$, intensity-windowed to range $[-200, 300]$ in Hounsfield units, and finally normalized to $[0,1]$.

\subsubsection{CNN baseline}\label{sec:unet}
As a \ac{cnn} baseline for semantic segmentation, we employ the U-Net \citep{Ronneberger2015UNet} with its extension of residual units (\texttt{MONAI} implementation). 
Each U-Net level employs the same number of channels as the corresponding MetaFormerS12 stage (64, 128, 320, 512).
To allow the U-Net to have a sufficiently large receptive field, we deploy an additional fifth stage with 1024 channels.
We construct U-Nets with $K\in\{3,5,7\}$ and additionally $K=9$ for the TIGER dataset because of its larger image size.
We match the U-Nets' number of parameters to be as close as possible to the corresponding MetaFormer with convolutional token mixers by varying the number of residual units at each stage.
To make sure, each stage is capable of extracting features, we set a minimum of one residual unit.\par
Since the MetaFormer’s first patch embedding uses a stride of four, its predicted segmentation map is coarser and must be upsampled accordingly (Sec.~\ref{sec:seg_architecture}).
For better comparison, we introduce a second U-Net variant denoted as \emph{UNet@PatchEmb} applied to a four times downsampled image by the same layer as the MetaFormer's first patch embedding.
This version omits the fifth U-Net stage to match the MetaFormer’s receptive field and adjusts the number of residual units to obtain a comparable parameter count.
Like in the MetaFormer, the predicted segmentation mask is upsampled by a factor of four to match the input resolution.

\subsubsection{Foundation Model Baseline}
As described in Sec.~\ref{sec:foundation_model}, we include RAD-DINO as a foundation model baseline.
With its \ac{vit} encoder frozen, we follow \cite{oquab2023dinov2} by extracting the $37\times37$ patch tokens from the last four layers as input to the SegFormer decoder (cf. Sec.~\ref{sec:seg_architecture}).
The \ac{vit} does not produce a hierarchical feature pyramid as in the other encoders, but keeps the number of channels and spatial dimensionality the same.
Here, all extracted patch tokens have a fixed dimensionality of $768$.
Consequently, with each stage projected to $C = 256$ within the SegFormer decoder, the amount of trainable parameters doubled with $1.1\,\mathrm{M}$ compared to MetaFormer-based variants (cf. Fig.~\ref{fig:architetcure}).

\subsection{Ranking} \label{sec:ranking}
To determine the overall best performing token mixer in each setting, we implement a ranking scheme based on the well-established challenges Medical Decathlon \citep{antonelli2022medicaldecathlon} and Learn2Reg \citep{hering2022learn2reg}.
Intuitively, every token mixer with a specific kernel size can be considered an individual challenge submission.
On each dataset, token mixers are compared against each other and earn a point, for each comparison they significantly won.
The results are sorted and token mixers are given scores from the range $[0.1, 1]$ per dataset.
In cases of ties, they share the average rank score.
Finally, an aggregated global rank is computed by the geometric mean across datasets.
Since for classification tasks, there is no instance-level performance with the \ac{auc} metric, we used bootstrapping of the test data with $5\,k$ repeats.
With it, we estimate the two-sided $95\,\%$ confidence interval of the difference in \ac{auc} to determine the significance of a token mixers' comparison.
For segmentation, we utilize the Wilcoxon signed-rank test with $p<0.05$ on \ac{dsc}.

\section{Results and Discussion}
\subsection{Classification}
\begin{table*}
    \centering
    \caption{\acl{auc} for different token mixers $T$ across datasets. The first meta row holds our models trained from scratch ($[T,T,T,T]$). The second utilizes ImageNet-pretrained weights for the channel-MLPs and only exchanges token mixers in the last two stages ($[P,P,T,T]$). Number of trainable parameters (\#P) is given in millions. See Sec.~\ref{sec:ranking} for ranking schema details. Refer to Fig.~\ref{fig:2p2t_uplift} for the impact of pre-training.\\ \footnotesize{* = reduced learning rate and doubled epochs; + = $\frac{\text{batch size}}{4}$ but gradient accumulation of 4; $\dagger$ disabled compilation (cf. Sec.~\ref{sec:training_instability}).}}
    \resizebox{!}{.455\textheight}{
    \begin{tabular}{r|lc|cccc|cc}
    \toprule
    & \multicolumn{2}{r|}{Dataset}  & Path & Derma & Pneumonia & OrganS & \\
    & \multicolumn{2}{r|}{(\# training images)} & ($90\,\mathrm{k}$) & ($7\,\mathrm{k}$) & ($6\,\mathrm{k}$) &($14\,\mathrm{k}$) & \\
    \cmidrule{2-3} \cmidrule{4-9} 
    & Token Mixer $T$ & Kernel $K$ & \multicolumn{4}{c|}{MedMNIST} & Rank & \#P\,[M] \\ 
    \midrule
    \multirow{14}{*}{\rotatebox[origin=c]{90}{$[T, T, T, T]$ from scratch}}&\multirow[c]{3}{*}{pooling} & 3 & 0.9758 & 0.8883 & \underline{0.9921} & 0.9617 & \underline{0.666} & 11.4\\
     && 5 & 0.9799 & 0.8911 & 0.9832 & 0.9375 & 0.429 & 11.4\\
     && 7 & 0.9829 & 0.8795 & 0.9814 & 0.9420 & 0.493 & 11.4\\
     \cmidrule{2-9}
    &\multirow[c]{3}{*}{conv} & 3 & 0.9834 & 0.8675 & 0.9761 & 0.9534 & 0.505 & 22.0\\
     && 5 & \underline{0.9878} & 0.8855 & 0.9807 & 0.9475 & 0.657 & 40.9\\
     && 7 & \textbf{0.9887} & 0.8628 & \textbf{0.9925} & 0.9503 & 0.622 & 69.2\\
     \cmidrule{2-9}
    &\multirow[c]{3}{*}{grouped conv} & 3 & 0.9803 & 0.8867 & 0.9886 & 0.9535 & \textbf{0.731} & 11.4\\
     && 5 & 0.9821 & 0.8746 & 0.9863 & 0.9410 & 0.483 & 11.5\\
     && 7 & 0.9798 & 0.8681 & 0.9878 & 0.9440 & 0.466 & 11.6\\
     \cmidrule{2-9}
    &\multirow[c]{3}{*}{$\text{local attn}^*$} & 3 & $\text{0.9701}^{\dagger}$ & 0.8869 & $\text{0.9739}^{\dagger}$ & 0.9384 & 0.349 & 16.1\\
     && 5 & $\text{0.9726}^{\dagger}$ & 0.8538 & 0.9780 & $\text{0.9274}^{\dagger}$ & 0.189 & 16.1\\
     && 7 & $\text{0.9782}^{\dagger}$ & 0.8458 & 0.9803 & $\text{0.9326}^{\dagger}$ & 0.28 & 16.1\\
     \cmidrule{2-9}
    &$\text{global attn}^+$ & - & 0.9802 & \textbf{0.9174} & 0.9493 & \textbf{0.9679} & 0.484 & 16.5\\
    \cmidrule{2-9}
    &random & - & 0.9805 & \underline{0.9024} & 0.9863 & 0.9461 & 0.588 & 11.4\\
    \cmidrule{2-9}
    &identity & 1 & 0.9756 & 0.8784 & 0.9883 & \underline{0.9626} & 0.528 & 11.4\\
    \midrule
    \multirow{18}{*}{\rotatebox[origin=c]{90}{$[P, P, T, T]$ pretrained channel-MLPs}}&\multirow[c]{3}{*}{pooling $P$} & 3 & 0.9596 & 0.9064 & 0.9830 & 0.9536 & 0.367 & 11.4\\
     && 5 & 0.9651 & 0.9191 & 0.9883 & \underline{0.9618} & 0.5 & 11.4\\
     && 7 & 0.9750 & 0.9198 & 0.9793 & 0.9475 & 0.479 & 11.4\\
     \cmidrule{2-9}
    &\multirow[c]{3}{*}{conv} & 3 & 0.9564 & 0.9127 & 0.9832 & 0.9545 & 0.367 & 21.7\\
     && 5 & 0.9678 & 0.9132 & 0.9861 & 0.9609 & 0.51 & 39.9\\
     && 7 & 0.9721 & 0.9143 & 0.9880 & 0.9567 & 0.523 & 67.2\\
     \cmidrule{2-9}
    &\multirow[c]{3}{*}{grouped conv} & 3 & 0.9620 & 0.9000 & 0.9906 & 0.9440 & 0.414 & 11.4\\
     && 5 & 0.9641 & 0.9046 & \underline{0.9918} & 0.9545 & 0.5 & 11.5\\
     && 7 & \underline{0.9800} & 0.9014 & 0.9864 & 0.9438  & 0.441 & 11.6\\
     \cmidrule{2-9}
    &\multirow[c]{3}{*}{local attn} & 3 & 0.9745 & 0.9097 & 0.9801 & 0.9489 & 0.479 & 16.0\\
     && 5 & 0.9742 & 0.9043 & \textbf{0.9935} & 0.9568 & \textbf{0.668} & 16.0\\
     && 7 & 0.9736 & 0.9197 & 0.9755 & 0.9554 & 0.523 & 16.0\\
     \cmidrule{2-9}
    &\multirow[c]{3}{6em}{local attn (warm start)} & 3 & 0.9753 & 0.9030 & 0.9814 & 0.9582 & 0.588 & 16.0\\
     && 5 & 0.9729 & 0.9032 & 0.9913 & 0.9449 & 0.475 & 16.0\\
     && 7 & 0.9796 & 0.9028 & 0.9894 & \textbf{0.9624} & \underline{0.641} & 16.0\\
     \cmidrule{2-9}
    &global attn & - & 0.9799 & 0.9046 & 0.9887 & 0.9512 & 0.614 & 16.0\\
    &(warm start) & - & \textbf{0.9847} & 0.9067 & 0.9915 & 0.9314 & 0.451 & 16.0\\
    \cmidrule{2-9}
    &random & - & 0.9744 & \underline{0.9425} & 0.9885 & 0.9404 & 0.458 & 11.4\\
    \cmidrule{2-9}
    &identity & 1 & 0.9511 & \textbf{0.9646} & 0.9911 & 0.9574 & 0.495 & 11.4\\
    \midrule
    \multirow{4}{*}{\rotatebox[origin=c]{90}{ResNet18}}&\multirow[c]{3}{*}{from scratch} & 3 & 0.9735 & 0.9135 & 0.8885 & \textbf{0.9627} && 11.2\\
    & & 5 & 0.9735 & 0.8772 & 0.9833 & 0.9558 && 30.7\\
    & & 7 & 0.9772 & 0.8639 & \textbf{0.9868} & 0.9378 && 60.0\\
    \cmidrule{2-9}
    & pretrained & 3 & \textbf{0.9871} & \textbf{0.9476} & 0.9390 & 0.9558 && 11.2 \\
    \midrule
    &RAD-DINO & - & 0.9744 & 0.8164 & 0.9947 & 0.9219 && 2.1\\
    \bottomrule
\end{tabular}}
    \label{tab:classification}
\end{table*}

\begin{figure*}
    \centering
    \includegraphics[width=\linewidth]{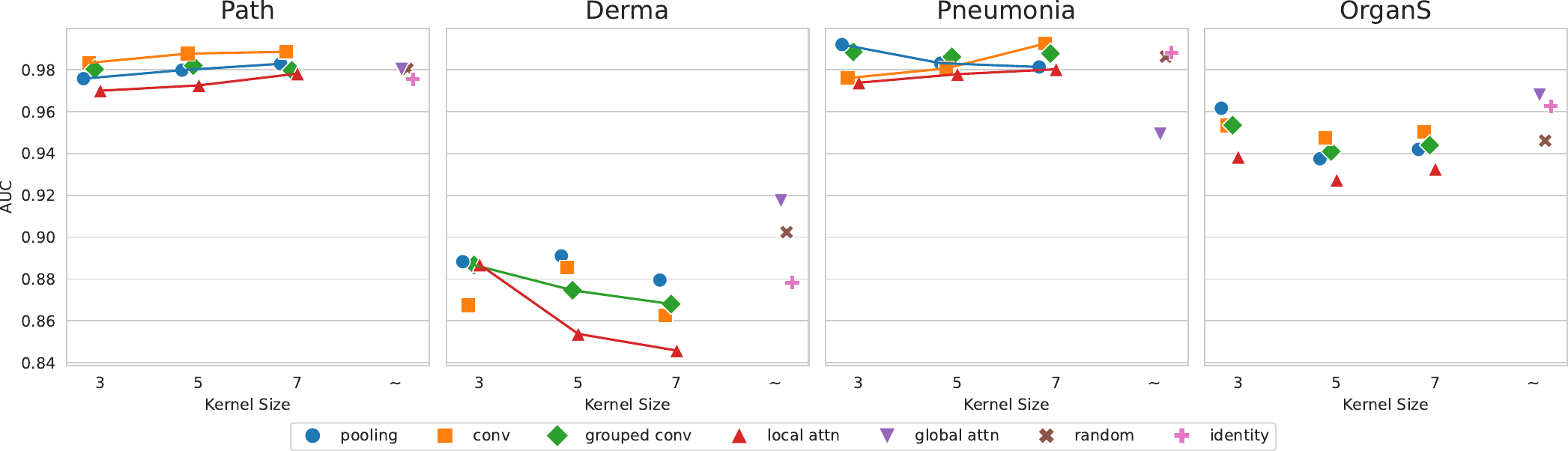}
    \caption{\acl{auc} for MetaFormer trained from scratch with different token mixers (architecture signature $[T, T, T, T]$). Line plots indicate results with a monotonic change across kernel sizes.}
    \label{fig:4t_scratch}
\end{figure*}

\subsubsection{Dataset-specific Performance of Token Mixers}\label{sec:classification_4t}
Fig.~\ref{fig:4t_scratch} and Tab.~\ref{tab:classification} (meta row $[T, T, T, T]$) show the \acf{auc} of MetaFormers trained from scratch utilizing the specified token mixer at each four stages.\par
On PathMNIST, which consists of pathology slide images where a larger spatial context is advantageous, we observe a positive effect from larger kernel size on the \ac{auc}, while the identity mapping yields lower performance.
Since PathMNIST provides $90\,\mathrm{k}$ training images, it allows for the effective optimization of models with many parameters.
This is reflected in the performance of the convolutional token mixers, which achieve the overall best results with a kernel size of 7.
This configuration adds the highest amount of trainable parameters with a large margin ($57.8\,\mathrm{M}$ parameters on top of the base $11.4\,\mathrm{M}$).\par
On DermaMNIST, token mixers with a global receptive field show the best performance with global attention being the overall best, followed by random token mixing.
For the other token mixers, larger kernels and increased parameter counts appear to hinder generalization.\par
In the chest X-rays of the PneumoniaMNIST dataset, pneumonia-infected tissue can span the entire lung, so a larger receptive field should be beneficial.
The result for convolution is consistent with this hypothesis.
However, for pooling and grouped convolution, different kernel sizes yield comparable \ac{auc}.
Moreover, token mixers with the largest receptive fields do not perform best; in fact, global attention performs worst among all token mixers on this task.
Nevertheless, as all token mixers (except global attention) achieve an \ac{auc} close to the optimum of 1.0, pneumonia detection can be regarded as a relatively easy classification task.
Hence, all models successfully learn discriminative features and, thus, the effect of an increased capacity (more parameters or larger kernels) is less noticeable.\par
For OrganSMNIST, which involves organ classification in abdominal \ac{ct} slices, no clear trend emerges regarding neither the optimal kernel size nor choice of token mixing.
The highest \ac{auc} is reached by global attention, followed by no token mixing (identity).\par
When comparing the correlation between kernel size and performance across datasets, one might observe a trend:
For datasets such as DermaMNIST, which typically contain a single, localized object per image (cf. Fig.~\ref{fig_ap:medmnist_montages} in appendix for image examples), kernel-based token mixing tends to favour smaller kernels.
Conversely, when the classification task requires capturing broader spatial context as in PathMNIST (pathology slides) or PneumoniaMNIST (chest radiographs, where infected tissue can span the entire lung) -- token mixers with a larger receptive field provide some advantage.

\subsubsection{Global Ranking of Token Mixers} \label{sec:classification_ranking}
Grouped convolution with a kernel size of $K=3$ obtains the overall best rank (0.731), followed by pooling with $K=3$ (0.666).
To make a more general statement, we averaged the rankings over the different token mixers and kernel sizes (see right column in Tab.~\ref{tab:classification}).
We observe that less complex token mixers perform better on average, putting convolution (0.594) first, followed by grouped convolution (0.56), pooling (0.53) and with a large margin local attention (0.273).
Moreover, smaller kernel sizes appear to be favourable, with $K=3$ ranking 0.563 on average and 5 and 7 comparable (0.4692/0.4653).
So in general, less complex token mixers with a smaller kernel sizes perform the best on average across different data sets.
See appendix Tab.~\ref{tab_ap:classification_ranking} for absolute and relative ranking scores for each dataset.
The noticeably lower ranks of attention-based token mixers with local on average of 0.273 and global with 0.484 suggest them less suitable when training from scratch on relatively small datasets.
However,  due to local attention's training instabilities (see Sec.~\ref{sec:training_instability}), we had to reduce the learning rate to $1e^{-4}$.
Although, we doubled the number of epochs for a comparable overall optimization budget, the reported \acp{auc} could still be underestimating local attention's potential.
The similar average rank of grouped convolution (0.56) and convolution (0.594) suggests that the channel cross-talk lacked by grouped convolution can be effectively compensated by the channel-MLP within the MetaFormer block.
Consequently, grouped convolution should be preferred, as it offers lower runtime complexity and reduced parameter count regarding the number of channels (cf. Tab.~\ref{tab:token_mixer_complexity}).
No token mixing (identity) reaches a competitive place of 6/15 (rank 0.528), which is consistent with the findings of \cite{yu2023metaformer}.
This raises whether additional token mixing is necessary at all, as the convolution (without any non-linearity) used for patch embedding at the beginning of each stage may already provide sufficient token mixing.
Another contributing factor could be the MetaFormer’s extension for global prediction: the final feature map is collapsed via \acf{gap} into a single vector (cf. Fig.~\ref{fig:architetcure}), which is essentially a global token mixing operation.
This might be sufficient token mixing for some global prediction tasks.
Furthermore, because each MetaFormer block includes skip connections (cf. Fig.~\ref{fig:metaformerblock}), the token mixing by \ac{gap} may propagate transitively through all stages of the network.

\subsubsection{Transfer Learning from ImageNet}\label{sec:classification_2p2t}
Another factor that may explain the poor performance of local self-attention, which ranked last in the previous experiment, is its slow convergence and high data requirements as reported in \cite{lu2022ViTonSmallData}.
To mitigate this limitation, we experiment with initializing our models with publicly available ImageNet-pretrained weights.
However, since pretrained weights are not available for all our deployed token mixers, we adopt a hybrid MetaFormer variant that uses pooling in the first two stages and global attention in the last two \citep{yu2022metaformeractuallyneedvision}.
To avoid a drastic domain shift, we keep pooling as token mixer of the first two stages (introducing no token mixer-parameters) and replace only the global attention in the last two stages with our token mixers.
This architecture signature is denoted as $[P, P, T, T]$ (see the corresponding meta row in Tab.~\ref{tab:classification}).
Because replacing the token mixer does not modify the channel-\ac{mlp}, we can leverage pretrained weights for all stages.
For self-attention (global and local), we differentiate between two variants:
\emph{First}, transfer the pretrained attention weight matrices $\mathbf{W}_{\{k,v,q,u\}}$ (cf. Sec.~\ref{sec:self-attention}) calling it “warm start” and \emph{second}, initializing their parameters randomly.
Since the architecture signatures between the two experiments differs ($[T, T, T, T]$ vs. $[P, P, T, T]$), we also train a $[P,P,T,T]$ variant from scratch (results shown in appendix Tab.~\ref{tab_ap:2p2t_scratch}), allowing us to measure the impact of pretrained weights.\par
\begin{figure*}
     \centering
     \begin{subfigure}{\textwidth}
         \centering
         \includegraphics[width=\linewidth]{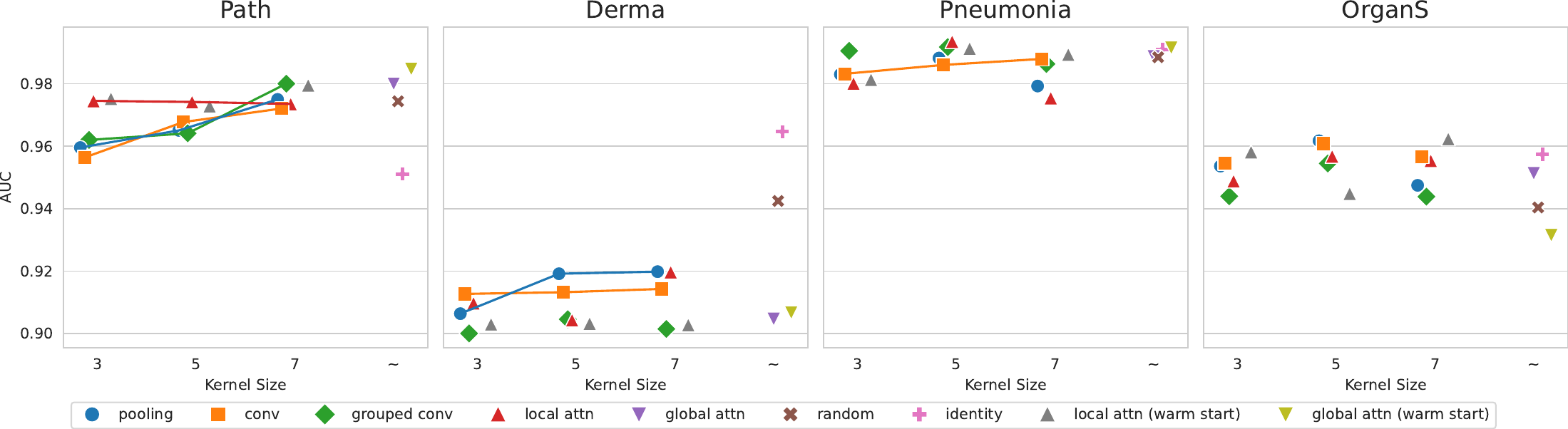}
         \caption{\acl{auc} for the MetaFormer using pretrained ImageNet weights utilizing different token mixers in the last two stages.}
         \label{fig:2p2t_pretrained}
     \end{subfigure}
     \begin{subfigure}{\textwidth}
         \centering
         \includegraphics[width=\linewidth]{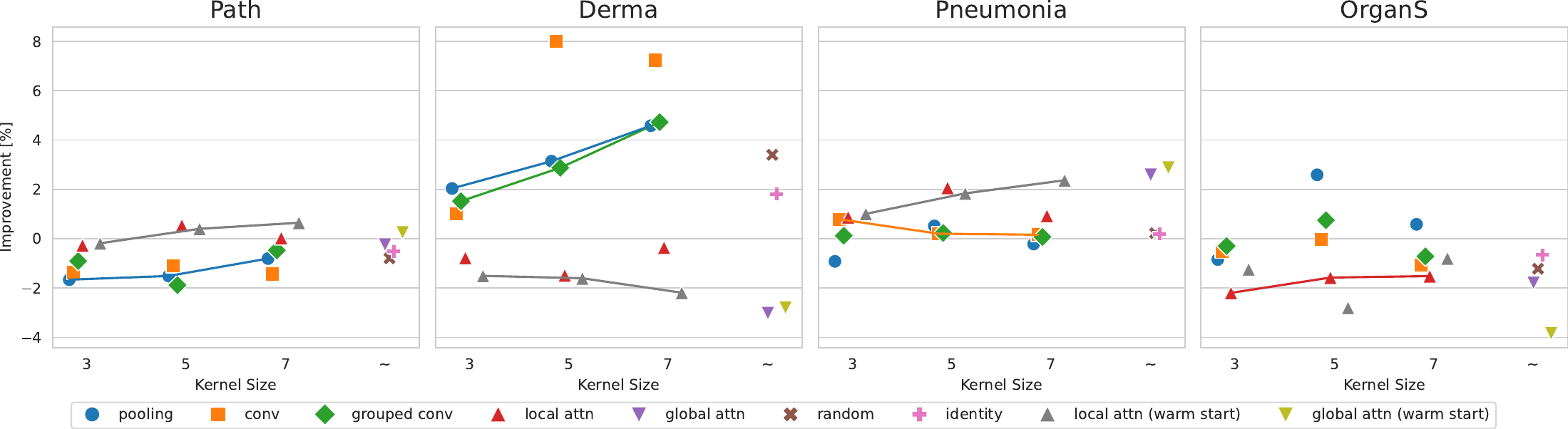}
         \caption{Relative improvement from utilizing pretrained weights compared to the variant trained from scratch.}
         \label{fig:2p2t_uplift}
     \end{subfigure}
        \caption{Performance of the MetaFormer using pretrained ImageNet weights for its channel-MLPs and different token mixers in the last two stages (architecture signature $[P,P,T,T]$). Line plots indicate results with a monotonic change across kernel sizes.}
        \label{fig:auc_results}
\end{figure*}

 
Fig.~\ref{fig:2p2t_pretrained} shows the absolute performance in \ac{auc} of the $[P,P,T,T]$ architecture utilizing pretrained weights, and Fig.~\ref{fig:2p2t_uplift} shows the relative improvement when compared to its counterpart trained from scratch.
Experiments on ImageWoof (appendix Sec.~\ref{sec_ap:imagewoof} in appendix) show that reusing global attention weights significantly boosts performance for global and local attention, particularly as local attention kernels increase in size and become more similar to global attention.
For non-attention-based token mixers, the improvement scales with similarity to global self-attention, with larger gains for parameterized convolutions than for pooling or identity token mixing.
In the medical setting, the impact of ImageNet pretrained weights is far less pronounced.
On PathMNIST, we observe the same positive correlation for larger context as we have in the $[T, T, T, T]$ variant trained from scratch.
However, the relative benefit of using pretrained weights is limited and is only noticeable for attention-based token mixers with kernel size $K \geq 5$.
On DermaMNIST, no token mixing (identity) outperforms any token mixing operation by a large margin (\ac{auc} of 0.9646, second best is random mixing at 0.9425.
Beyond its strong absolute performance, identity mapping does not benefit much from pretrained weights on DermaMNIST.
Attention-based mixers even perform worse when pretrained compared to training from scratch on this dataset.
In contrast, convolution- and pooling-based token mixers gain the most, with larger kernels showing greater improvements.
These improvements are the largest among the medical datasets.
We attribute this to the dataset’s domain -- colour microscope images of skin lesions, typically a single localised object -- which is closer to the ImageNet domain and suits pooling and convolutional layers.
Additionally, because DermaMNIST has the lowest \acp{auc} compared to the other datasets, it offers the greatest potential for performance improvement.
In PneumoniaMNIST, we find the same positive correlation between kernel size and absolute performance and a benefit of up to $2.5\%$ improvement when using pretrained weights for all token mixers except pooling.
As in the previous experiment, we do not observe any clear performance dependencies on the OrganSMNIST dataset.
Here, most token mixers yield worse performance when using pretrained ImageNet weights compared to training from scratch.
When averaging the relative improvements of all token mixers across the medical datasets, convolution achieves the largest gain ($+0.9\,\%$), while global attention with warm-start on average does not benefit from pretrained weights ($-0.8\,\%$).
These mixed results imply that ImageNet pretrained weights can only be applied to medical image classification with some caveat.
However, the mean improvement per kernel size $K$ further supports our interpretation, that moving closer to the receptive field of the original token mixer (global attention) helps narrow this domain gap, with $K=7$ yielding $+0.59\,\%$, $+0.56\,\%$ for $K=5$, and $-0.004\,\%$ with $K=3$.\par
When considering the global ranks, we find the first four ranks (19 entries in total, see Tab.~\ref{tab:classification}) holding attention-based token mixers. 
Local attention with kernel size $K=5$ and $K=7$ hold the first (0.668) and second (0.641) place, respectively.
We attribute the lower ranks of non-attention-based token mixers to the domain gap introduced by replacing the original attention operation, hindering the benefit from pretrained channel-MLPs.
However, the low rank (0.451) of global attention with warm-start initialization (15th place) suggests that the superior performance of local attention cannot be explained solely by its similarity to the pretrained global attention, but also by the presence of a local inductive bias.
Convolution, grouped convolution and pooling rank similarly, with an average rank of 0.466, 0.4517 and 0.449, respectively.
Interestingly, identity achieves a competitive rank of 0.495, making it the first non-attention-based token mixer in the ranking.
This suggests again that token mixing may be of secondary importance in a global prediction setting, where the final feature map is spatially averaged into a single vector before the classification head.
See appendix Tab.~\ref{tab_ap:classification_ranking} for absolute and relative ranking scores for each dataset.

\subsubsection{Training Instabilities of Flex Attention}\label{sec:training_instability}
We employ PyTorch's \texttt{flexattention} framework to implement the necessary attention pattern for local attention (cf. Eq. (\ref{eq:local_mha})).
The framework heavily relies on \texttt{torch.compile} to reduce both memory consumption and computational overhead.
By monitoring the per-layer gradient $L_2$-norm, we observe that compilation can induce gradient explosions within the flexattention module when training on large input resolutions.
For classification, this limit is at the spatial resolution of MetaFormer's first stage (in our setting $56\times56$).
To stabilize training, we experimented with gradient clipping, returning to full 32 bit precision (we usually utilize 16 bit \texttt{bfloat} for efficiency) and also prescaling $\mathbf{q}$ by $\frac{1}{\sqrt{D}}$ before its correlation with $\mathbf{k}$ to reduce the magnitude of values within the matrix multiplication (cf. Eq.~\ref{eq:local_mha}).
In the end, reducing the learning rate from $1e^{-3}$ to $1e^{-4}$ solved the issue for the most situations.
To keep the training budget comparable, we double the number of epochs (marked with * in Tab. \ref{tab:classification}).
In cases, where reducing the learning rate was not sufficient, we are forced to disable the compilation of flexattention (marked with $\dagger$ in Tab.~\ref{tab:classification}).
This massively increases the resource consumption from one hour training time with around $10\,\text{GB}$ VRAM at a batch size of 128 to 20 hours and $60\,\text{GB}$ at a batch size of 64.
To compensate for the reduced batch size, we employ gradient accumulation to simulate an update with the original batch size.
Interestingly, we do not encounter any issues when employing the MetaFormer as encoder for segmentation regardless of the larger input sizes up to $768\times768$ for the TIGER task.
We attribute this stability to the SegFormer decoder, which adds shortcuts to each MetaFormer stage (cf. Fig.~\ref{fig:architetcure}) and thereby shortens the effective gradient path.

\subsubsection{ResNet Baseline}
The ResNet18 baseline trained from scratch gets outperformed by its MetaFormer counterpart (employing convolutional token mixer in $[T,T,T,T]$) on PathMNIST (by $1.2\,\%$ \ac{auc}) and PneumoniaMNIST (by $3\,\%$ \ac{auc}) on average.
The same holds for grouped convolution with $0.5\,\%$ and $3.5\,\%$ \ac{auc}, respectively.
On the two more challenging medical datasets (DermaMNIST and OrganSMNIST), the ResNet18 performs on par with the MetaFormer.
Overall, we observe the ResNet18 following the same trends for different kernel sizes across datasets as we saw in the MetaFormer.\par
For the ResNet18 employing ImageNet pretrained weights, we find a similar pattern implying that ImageNet pretrained weights do not provide a substantial improvement on MedMNIST: using pretrained ImageNet weights yields the largest gains DermaMNIST ($+3.73\,\%$) and PathMNIST ($+1.01\,\%$), but results in a performance drop on PneumoniaMNIST ($-4.84\,\%$) and OrganSMNIST ($-0.69\,\%$).

\subsubsection{RAD-DINO Baseline}
While the consistently strong \ac{auc} scores indicate that RAD-DINO provides highly generalizable and discriminative features, it is still outperformed by the best end-to-end trained models.
An exception is observed on PneumoniaMNIST.
RAD-DINO was pretrained on large chest radiograph datasets such as CheXpert, ChestX-ray14, and PadChest, making it a particularly effective off-the-shelf model for pneumonia classification, achieving the highest overall \ac{auc} of $0.9947$.
Note that the PneumoniaMNIST dataset \citep{Kermany2018_pneumoniamnist} is not included in RAD-DINO pretraining data.

\subsection{Segmentation}
\begin{table*}
    \centering
    \caption{\acf{dsc} across three medical datasets for semantic segmentation. We exchange the token mixers at all four stages. All models are trained from scratch. See Sec.~\ref{sec:ranking} for ranking schema details. Number of trainable parameters (\#P) is given in millions.}
    \begin{tabular}{c|lc|ccc|cc}
    \toprule
     &\multicolumn{2}{r|}{Dataset (number of training images)}  & \multirow{2}{3.5em}{\centering JSRT (160)} & \multirow{2}{3.5em}{\centering GRAZ (43)} &  \multirow{2}{5em}{\centering TIGER@768² (256)} & \\
     \cmidrule{2-3}
     &TokenMixer $T$ & Kernel $K$   & & & & Rank & \#P[M] \\
     \midrule
    \multirow{18}{*}{\rotatebox[origin=c]{90}{$[T, T, T, T]$ from scratch}}&\multirow[c]{4}{*}{pooling} & 3 & 0.9468 & 0.8094 & \textbf{0.6125} & 0.614 & 11.9\\
     && 5 & 0.9463 & 0.8072 & 0.5555 & 0.49 & 11.9\\
     && 7 & 0.9450 & 0.7973 & 0.5580 & 0.418 & 11.9\\
     && 9 &  &  & 0.6019 & & 11.9\\
     \cmidrule{2-8}
    &\multirow[c]{4}{*}{conv} & 3 & 0.9486 & 0.8164 & \underline{0.6024} & 0.796 & 22.6 \\
     && 5 & \textbf{0.9510} & 0.8276 & 0.5622 & 0.729 & 41.4\\
     && 7 & \underline{0.9504} & \textbf{0.8330} & 0.5623 & 0.71 & 69.7\\
     && 9 &  &  & 0.5689 & & 107\\
     \cmidrule{2-8}
    &\multirow[c]{4}{*}{grouped conv} & 3 & 0.9500 & 0.8119 & 0.5791 & 0.625 & 12.0 \\
     && 5 & 0.9495 & 0.8319 & 0.5941 & \underline{0.836} & 12.0\\
     && 7 & \underline{0.9504} & \underline{0.8320} & 0.5711 & \textbf{0.843} & 12.1\\
     && 9 &  &  & 0.5893 & & 12.2\\
     \cmidrule{2-8}
    &\multirow[c]{4}{*}{local attn} & 3 & 0.9418 & 0.8165 & 0.5167 & 0.224 & 16.7\\
     && 5 & 0.9465 & 0.7957 & 0.5303 & 0.388 & 16.7\\
     && 7 & 0.9449 & 0.7927 & 0.5374 & 0.309 & 16.7\\
     && 9 &  &  & 0.5511 & & 16.7\\
     \cmidrule{2-8}
    &global attn & - & 0.9443 & 0.7562 & DNS & 0.207 & 16.7\\
    \cmidrule{2-8}
    &random & - & 0.9472 & 0.7441 & 0.5196 & 0.286 & 11.9\\
    \cmidrule{2-8}
    &identity & 1 & 0.9458 & 0.7417 & 0.5358 & 0.215 & 11.9\\
    \midrule
    \multirow{8}{*}{\rotatebox[origin=c]{90}{CNN baseline}}&\multirow[c]{4}{*}{UNet} & 3 & \textbf{0.9552} & \textbf{0.8478} & 0.5666 & & 28.4\\
     && 5 & 0.9499 & 0.8258 & 0.5458 & & 42.1\\
     && 7 & 0.9505 & 0.8318 & 0.5533 & & 82.1\\
     && 9 &  &  & 0.5544 & & 135\\
     \cmidrule{2-8}
    &\multirow[c]{4}{*}{UNet@PatchEmb} & 3 & 0.9357 & 0.7738 & 0.5735 & & 21.2\\
     && 5 & 0.9342 & 0.7864 & 0.5981 & & 39.4\\
     && 7 & 0.9335 & 0.7629 & \textbf{0.6047} & & 77.1\\
     && 9 &  &  & 0.5639 & & 96.6\\
     \midrule
     & RAD-DINO & - & 0.9424 & 0.5216 & 0.5734 & & 1.1\\
     \bottomrule
    \end{tabular}
    \label{tab:segmentation}
\end{table*}

\begin{figure*}
    \centering
    \includegraphics[width=.75\linewidth]{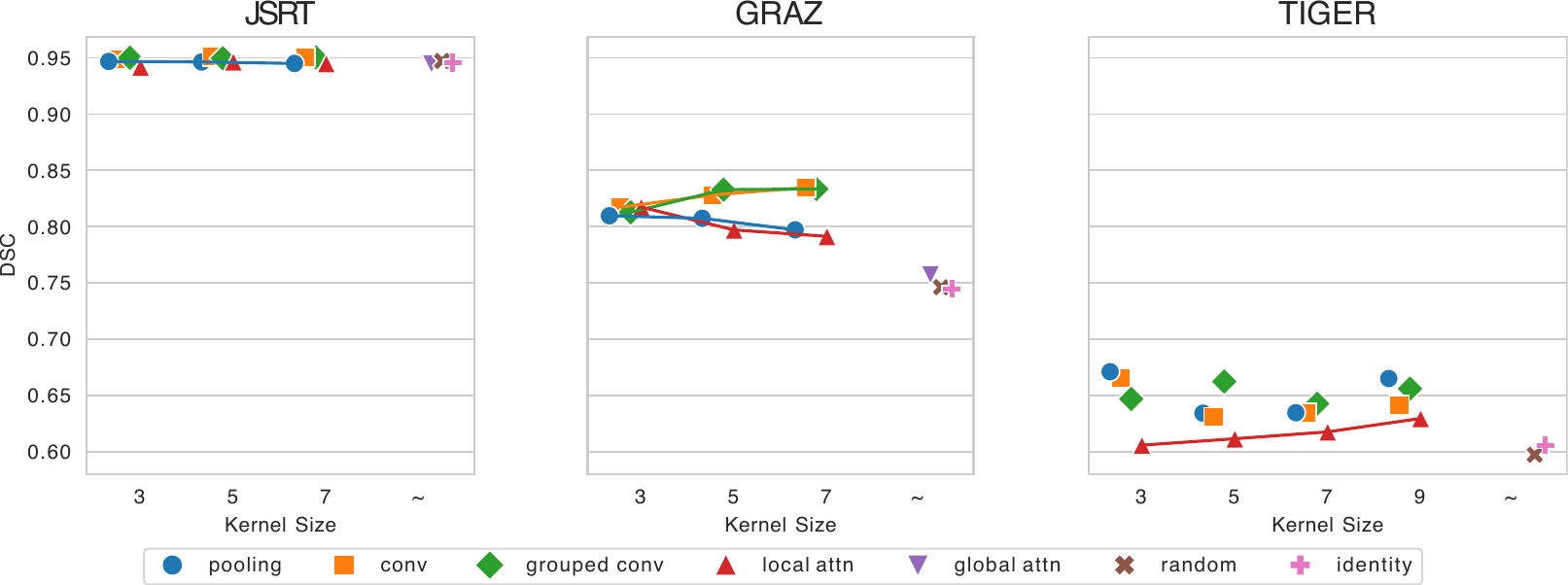}
    \caption{\acl{dsc} across three medical datasets for semantic segmentation, investigating into different token mixer and kernel sizes. Line plots indicate performance with a monotonic change across kernel sizes.}
    \label{fig:seg_dsc}
\end{figure*}

\subsubsection{Dataset-specific Performance of Token Mixers}
For semantic segmentation as a dense prediction task, we employ the MetaFormer as image encoder with different token mixers at every stage, hence having an architecture signature of $[T, T, T, T]$.
We extend its architecture with a lightweight \acs{mlp}-only decoder of the SegFormer \citep{xie2021segformer} (see Fig.~\ref{fig:architetcure} and Sec. \ref{sec:seg_architecture} for details).
Tab.~\ref{tab:segmentation} and Fig.~\ref{fig:seg_dsc} show the average \acf{dsc} across three datasets for different token mixers.
Exemplary qualitative segmentation results are shown in Fig.~\ref{fig_ap:qualitative_segmentation_results} in the appendix.\par
Segmentation of the lungs, heart, and clavicles in JSRT chest X-rays constitutes a relatively straightforward task, as all evaluated models achieve a \acf{dsc} of at least 0.94.
As a result, this dataset does not provide informative distinctions in performance between different token mixers or kernel sizes.\par
Next, we consider the GRAZ dataset, which involves segmenting 17 bones in paediatric wrist radiographs and presents challenges such as small, overlapping carpal bones.
The results highlight the importance of an inductive bias with strong locality, as convolutional token mixers with the kernel size $K>3$ outperform pooling- and attention-based alternatives.
Here, larger kernels further improve performance, with $K=7$ yielding an increase of nearly 2 \ac{dsc} percent points compared to $K=3$ ($83.3\,\%$ vs. $81.6\,\%$).
Global self-attention and random mixing perform worst, indicating that while a large receptive field is available, high local inductive bias is critical for this task.
However, the overall performance of attention-based token mixers may be constrained by the limited dataset size: 43 training images may be insufficient to reliably learn attention patterns.\par
We use the opportunity provided by the large histology slides of the TIGER dataset to employ an additional kernel size $K=9$ to evaluate if and to which extent a larger receptive field is beneficial for such tasks.
As it is common for histology tasks with high-resolution images, we settle for a patch-based approach using patches of size $768\times768$ (see Sec.~\ref{sec:seg_architecture} for details).
While we observe a positive correlation between local self-attention performance and a larger receptive field size, no such clear trend exists for the other token mixers.
It would have been informative to evaluate whether this trend extends to the global receptive field by global attention; however, the sequence length resulting from our patch size ($768^2 = 589\,824$) exceeds the capacity of even our largest available GPU (NVIDIA H200 with $144\,\text{GB}$ VRAM) regardless of employing gradient accumulation reducing the batch size down to 1 and employing gradient checkpointing after each MetaFormer block.
As a non-parametric alternative with a global receptive field, we employ random token mixing.
However, we find it performing worse with a \ac{dsc} of $51.2\,\%$ and thus indicate that a global receptive field is not beneficial for non-parametric token mixing.
Although no consistent pattern emerges for the other token mixers, all of them outperform even the best-performing local attention with $K=9$.\par

\subsubsection{Global Ranking of Token Mixers}
The global rank scores in Tab. \ref{fig:seg_dsc} (right column) suggest that dense prediction tasks require explicit token mixing at every stage, which was not the case for global prediction tasks (cf. Sec.~\ref{sec:classification_ranking}).
We find grouped convolution with a large kernel size $K=7$ yielding the highest global rank (0.843) followed by its $K=5$ variant (0.836).
However, we cannot observe the pattern of larger kernels yielding better ranks for the other token mixers.
Furthermore, considering the locality of the ranked token mixers (averaged global rank over kernel sizes), from the high locality of grouped/standard convolution (0.768/0.745) to pooling (0.507), local attention (0.307), and finally global attention (0.207), it is evident that a strong local inductive bias is crucial for segmentation in addition to the token mixing itself.
This lack of inductive bias hinders the performance of global attention, especially for the small datasets at hand.
The same applies for random token mixing and the identity operation, which rank similarly low as global attention does (0.286/0.215).
As already observed in the classification experiments, using standard convolution (rank 0.745) as the token mixer introduces unnecessary runtime and parameters, since the channel cross-talk that grouped convolution (rank 0.768) would lack is already provided by the MetaFormer block’s channel-\ac{mlp}.
See appendix Tab.~\ref{tab_ap:seg_ranking} for absolute and relative ranking scores for each dataset.

\subsubsection{U-Net Baselines}
For the \ac{cnn} baseline results, see Tab.~\ref{tab:segmentation}.
The U-Net exhibits a negative correlation between performance and kernel size, suggesting overfitting due to high number of parameters at larger kernels on our limited medical datasets.
However, the variant with kernel size $K=3$ outperforms the best-performing MetaFormer on the JSRT and GRAZ datasets.
Since the initial patch embedding in the MetaFormer downsamples the input by a factor of four, the predicted segmentation mask is coarser compared to the U-Net.
For a more fair comparison, we also evaluate a U-Net applied to the same downsampled patch embedding (UNet@PatchEmb, see Sec.~\ref{sec:seg_architecture}).
In this setting, U-Net performs noticeably worse than the MetaFormer with convolutional token mixers.
While we observe a positive correlation with larger kernels on the TIGER dataset for UNet@PatchEmb with kernel size $K\leq7$, the high number of parameters ($96.6\,\text{M}$) associated with $K=9$ may be difficult to train effectively given the limited dataset size.
Note that for each of the 256 TIGER train images, multiple $(768\times768)$-sized patches can be extracted, increasing the effective dataset size.

\subsubsection{RAD-DINO Baseline}
On the JSRT and TIGER datasets, the SegFormer decoder operating on RAD-DINO features provides a strong baseline, achieving \ac{dsc} scores of 0.9424 and 0.5734, respectively, placing it in the midrange compared to other models.
However, a key limitation of using RAD-DINO’s \ac{vit} as the encoder is the absence of a hierarchical feature pyramid and the relatively low feature resolution ($37\times37$), which necessitates upsampling the logits by up to a factor of ten.
As a consequence, the model struggles to segment small anatomical structures, such as the carpal bones in the GRAZ setting, resulting in a \ac{dsc} of only 0.5216 (-0.3114 compared to the best-performing MetaFormer variant).

\subsection{Token Mixing in 3D Domains}
As medical imaging commonly involves volumetric data (e.g., CT or MRI), we extend the MetaFormer and its token mixers to the 3D domain.
This setting is particularly compelling, as foundation models have rapidly advanced 2D image analysis \citep{liang2025vfm,ma2024medsam,chen2024uni,perez2025raddino}, while their 3D counterparts are only beginning to emerge \cite{xu2025dino3d}.
Consequently, an encoder that supports interchangeable token mixers with varying spatial complexity offers strong flexibility for both global and dense prediction tasks in 3D (e.g., large receptive fields vs. fine-grained feature abstraction).
In this section, we present one example each for 3D global and 3D dense prediction tasks.
Due to memory constraints and the cubically scaling parameter count for, e.g., convolution, we limit our experiments to kernel sizes of $K\in\{3, 5\}$.
Tab.~\ref{tab:3d_task} represents the performance for classification in \ac{auc} and for segmentation in \ac{dsc}.

\subsubsection{Lung Nodule Classification}
Local attention achieves the highest \ac{auc} with a kernel size of $K=3$, followed by grouped convolution ($K=5$), random mixing, and pooling ($K=3$).
Increasing the kernel size does not lead to consistent improvements.
In particular, pooling, and local attention show slight performance drops for larger kernels.
Additionally, global token mixing does not improve performance for 3D lung nodule classification.
Overall, these observations indicate that local and structured token mixing strategies tend to be more effective, while larger receptive fields or global interactions do not provide clear benefits in this setting.
For the \ac{cnn} baselines, ResNet18 with a kernel size of $K=5$ outperforms its $K=3$ counterpart, but ranks below most MetaFormer variants.
Considering both performance (0.8923 vs. 0.8494 \ac{auc}) and parameter count ($20\,\textsc{M}$ vs. $152\,\textsc{M}$), the MetaFormer with local self-attention provides a more efficient and better-performing alternative to the 3D ResNet18.

\subsubsection{Abdominal Organ Segmentation in CT}
We observe that the local inductive bias of convolution is beneficial for segmenting abdominal organ structures in CT.
This is consistent with our findings in the 2D domain.
However, the narrow performance margins between alternative token mixers prevent us from definitively concluding that strict locality is the primary driver of performance in this setting.
For instance, local self-attention underperforms relative to global self-attention, and random token mixing yields results comparable to grouped convolutions.
Furthermore, we do not observe a clear or consistent trend regarding the two kernel sizes tested. 
While a smaller kernel ($K=3$) is beneficial for pooling, convolution, and local attention, this is not the case for grouped convolution.
Collectively, these results suggest that the model benefits more from the specific local inductive bias offered by learnable parameters in standard convolutions than from an expanded receptive field.
Regarding the trade-off between performance and efficiency, pooling with $K=3$ offers a compelling alternative: it achieves a \ac{dsc} of $0.9405$ compared to the $0.9444$ achieved by standard convolution, while utilizing only $15.8\,\textsc{M}$ parameters versus $47.1\,\textsc{M}$.
The UNet provides a strong baseline, outperforming all but the top-performing MetaFormer variant (convolution with $K=3$).
Notably, even the smaller UNet configuration with $K=3$ remains significantly more parameter-heavy, requiring $84.2\,\textsc{M}$ parameters compared to $47.7\,\textsc{M}$.

\begin{table*}
    \centering
    \caption{Performance for different token mixers $T$ for lung nodule classification (Class) and abdomen organ segmentation (Seg) in 3D stated as \ac{auc} and \ac{dsc}, respectively. Number of trainable parameters per task (\#P) is given in millions.\\\footnotesize{$\dagger$ disabled compilation (cf. Sec.~\ref{sec:training_instability}).}}
    \begin{tabular}{r|lc|c|c|c}
    \toprule
    & \multicolumn{2}{r|}{Dataset}  & Nodule Class & Abdomen Seg \\
    & \multicolumn{2}{r|}{(\# training images)} & 1158 & 4675 & \#P\,[M] \\
    \cmidrule{1-3}
    & Token Mixer $T$ & Kernel $K$ & \acs{auc} & \acs{dsc} &  Class/Seg\\ 
    \midrule
    \multirow{14}{*}{\rotatebox[origin=c]{90}{$[T, T, T, T]$ from scratch}}&\multirow[c]{2}{*}{pooling} & 3 & 0.8685 & 0.9405 & 15.2/15.8\\
     && 5 & 0.8582 & 0.9387 & 15.2/15.8\\
     \cmidrule{2-6}
    &\multirow[c]{2}{*}{conv} & 3 & 0.7931 & \textbf{0.9444} & 47.1/47.7\\
     && 5 & \underline{0.8903} & \underline{0.9408} & 162/162.6\\
     \cmidrule{2-6}
    &\multirow[c]{2}{*}{grouped conv} & 3 & 0.8705 & 0.9379 & 15.3/15.9\\
     && 5 & 0.8704 & 0.9397 & 15.7/16.3\\
     \cmidrule{2-6}
    &\multirow[c]{2}{*}{local attn} & 3 & \textbf{0.8923} & 0.9295 & 20/20.6\\
     && 5 & $\text{0.8345}^\dagger$ & 0.9266 & 20/20.6\\
     \cmidrule{2-6}
    &global attn& - & 0.8646 & 0.9329 & 20/20.6\\
    \cmidrule{2-6}
    &random & - & 0.888 & 0.9355 & 15.2/15.8\\
    \cmidrule{2-6}
    &identity & 1 & 0.8586 & 0.9357 & 15.2/15.8\\
    \midrule
    \multirow{2}{*}{\rotatebox[origin=c]{90}{CNN}}&\multirow[c]{2}{*}{ResNet18/UNet} & 3 & 0.8114 & 0.9434 & 33.2/84.2\\
    & & 5 & 0.8494 & 0.9439 & 152/208\\
    \bottomrule
\end{tabular}
    \label{tab:3d_task}
\end{table*}

\section{Conclusion and Future Work}
In this work, we systematically compared different token-mixing operations -- pooling, (grouped) convolution, random mixing, local and global self-attention -- within the MetaFormer architecture across nine datasets, seven in the 2D and two in the 3D medical image domain.
We evaluated two task categories: \emph{global prediction}, using five image classification datasets, and \emph{dense prediction}, using four semantic segmentation datasets, comparing our results against a foundation model and \ac{cnn}-based baselines.\par
When comparing different kernel sizes for classification, we observe that datasets containing a single dominant object in the image (e.g., DermaMNIST) favour more locality when employing kernel-based token mixing and thus benefit from smaller kernel sizes.
In contrast, tasks with larger regions of interest, such as PathMNIST and PneumoniaMNIST, tend to profit from incorporating a broader spatial context through larger kernels.
Based on our ranking schema, we found simpler token mixers like grouped convolution and pooling favourable when training from scratch.
Besides their good performance, their low runtime complexity and small amount of trainable parameter (if any) make them a sensible choice.
Utilizing a fixed, random global token mixing provides competitive performance (particularly when training from scratch).
However, despite introducing no additional trainable parameters, the fixed mixing weights increase the model’s parameter count by approximately 21M (for the S12 variant at an input resolution of $224\times224$).
Due to this and its quadratic complexity, we view this approach primarily as a methodological baseline rather than a practical solution for real-world applications.
In contrast, even omitting token mixing entirely (using the identity operation) yields comparable classification performance.
We attribute this to MetaFormer's classification extension, which employs \acf{gap} on the final feature map, effectively providing a sufficient implicit token-mixing.
Our analysis of transferring ImageNet-pretrained channel-\acsp{mlp} shows that the domain gap to medical images leads to only inconsistent improvements across datasets.
Still, gains of up to $8\,\%$ with pre-trained weights for specific datasets and token-mixers suggest testing its applicability on a case-by-case basis.
Therefore, releasing pretrained models trained on a diverse set of medical datasets could provide a valuable contribution to the community in the future.\par
For segmentation tasks, we observe that a strong local inductive bias is crucial, making convolutional token mixers the most effective choice.
Here, grouped convolution achieves the highest rank across datasets while offering lower runtime complexity and fewer parameters than standard convolution.
We attribute this superior performance to the channel-\acp{mlp}, which already provide the channel cross-talk that gets omitted in grouped convolution.
Because our segmentation decoder (a lightweight SegFormer design \citep{xie2021segformer}) does not introduce token mixing, this must be provided within the MetaFormer.
Consequently, using no token mixing (identity) leads to inferior performance (second to last global rank).
We explicitly chose a lightweight decoder to emphasize the impact of different token mixers.
Future work might explore alternative decoder designs that incorporate varying degrees of token mixing like MetaSeg \citep{Kang2024MetaSeg}.\par
To address the central research question of this work -- how to mix tokens (Shaken or Stirred?) within the MetaFormer for medical image processing -- our findings suggest that low-complexity token mixers such as grouped convolution, pooling, or even the identity operation are sufficient for classification tasks.
We observe no major performance gains from using more complex mixers when training from scratch, confirming that the insights from MetaFormer studies on natural images~\citep{yu2023metaformer, yu2022metaformeractuallyneedvision} also generalize to the medical imaging domain.
For segmentation tasks, however, we find strong locality crucial and therefore recommend grouped convolution as the preferred token mixer.\par
While the included RAD-DINO foundation model delivers strong performance in both classification and segmentation tasks (particularly for pneumonia classification, due to its pretraining on numerous chest radiographs), it is ultimately outperformed by models trained end-to-end.
Since the medical domain also includes volumetric data (e.g., CT or MRI), we evaluate one example each for classification and segmentation.
Our 3D results confirm the 2D findings, with local token mixing outperforming approaches with large receptive fields, while the cubic complexity in 3D further favours simpler token mixers.
While our study has focused on classification and segmentation as representative global and dense prediction tasks, other applications such as object detection are also of importance in the medical domain.
Future work could explore using the MetaFormer as an image encoder within the \ac{detr} framework~\citep{carion2020detr}.
Since our work focuses on the token mixing within the encoder, another future direction could include investigation of the interplay between encoder and decoder token mixing, including more expressive decoder designs like MetaSeg \citep{Kang2024MetaSeg}, to better understand their joint impact on segmentation performance.


\acks{This research has been funded by the state of Schleswig-Holstein, Germany, Grant Number 22023005.}

%
\ethics{The work follows appropriate ethical standards in conducting research and writing the manuscript, following all applicable laws and regulations regarding treatment of animals or human subjects.}

\coi{We declare we don't have conflicts of interest.}

\data{All the data used in this work are publicly available and can be accessed through the cited sources. To reproduce the preprocessing steps and our results, please follow the instructions provided in our GitHub repository: \url{https://github.com/multimodallearning/MetaFormerMedImaging/tree/clean_code}.}

\section*{Declaration on Generative AI}
During the preparation of this work, the authors used ChatGPT to paraphrase and reword.
After using this tool/service, the authors reviewed and edited the content as needed and take full responsibility for the publication’s content.

\bibliography{ShakenOrStirred}


\clearpage
\appendix
\section{Overview of Appendix}
\begin{figure*}[h]
    \centering
    \begin{subfigure}{\textwidth}
        \centering
        \begin{subfigure}[t]{.49\linewidth}
            \includegraphics[width=\textwidth]{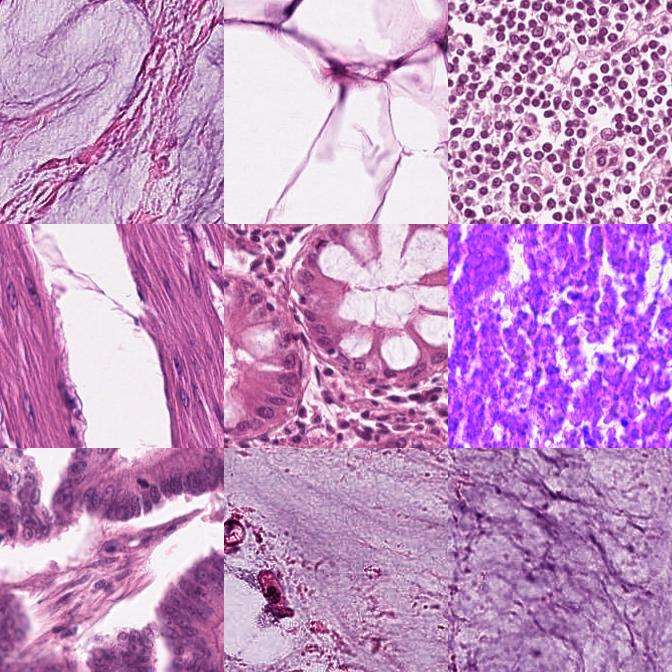}
            \caption{Path}
        \end{subfigure}\hfill
        \begin{subfigure}[t]{.49\textwidth} 
            \includegraphics[width=\textwidth]{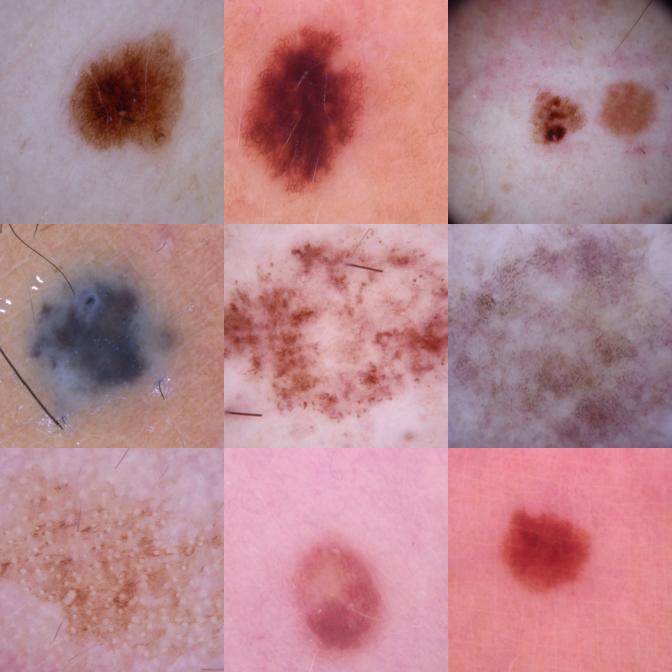}
            \caption{Derma}
        \end{subfigure}
    \end{subfigure}
    \begin{subfigure}{\textwidth}
        \centering
        \begin{subfigure}[t]{.49\linewidth}
            \includegraphics[width=\textwidth]{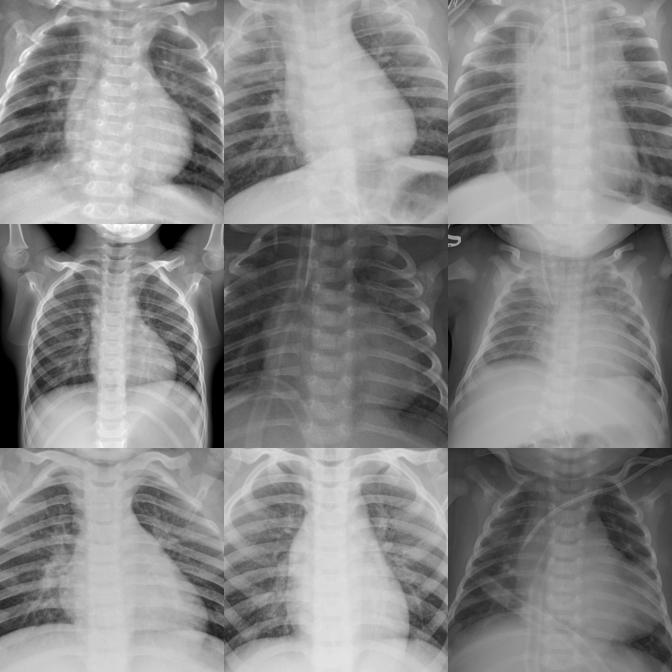}
            \caption{Pneumonia}
        \end{subfigure}\hfill
        \begin{subfigure}[t]{.49\textwidth} 
            \includegraphics[width=\textwidth]{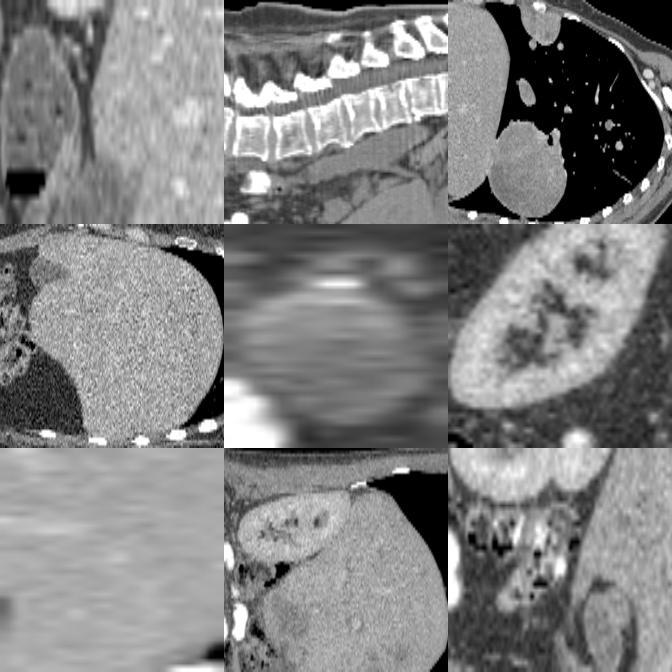}
            \caption{OrganS (s=sagittal view)}
        \end{subfigure}
    \end{subfigure}
    \caption{Nine examples of each 2D MedMNIST classification dataset used in our experiments (randomly sampled from the test split).}
    \label{fig_ap:medmnist_montages}
\end{figure*}

The appendix provides additional experiments on other MetaFormer's architecture S-variants (Sec.~\ref{sec_ap:s_variants}).
It also reports results on the natural image ImageWoof dataset, which are excluded from the main manuscript, as the focus is on medical applications of MetaFormer (see Sec.~\ref{sec_ap:imagewoof} for details).
Furthermore, we include example images from the 2D MedMNIST datasets (Fig.~\ref{fig_ap:medmnist_montages}) and qualitative results for the 2D segmentation tasks (Fig.~\ref{fig_ap:qualitative_segmentation_results}).
To provide additional insights into our 2D classification results  including ImageWoof), we include both accuracy and F1-score as supplementary metrics.
The F1-score might offer more intuition than the \ac{auc}, as it represents the harmonic mean of precision and recall, making it a more interpretable indicator of classification performance.
Nevertheless, since the \ac{auc} evaluates classifier performance across various confidence thresholds, we consider it the primary metric for discussing the classification benchmarks in our work.
In addition, we report the absolute and relative ranking scores of all token mixers across datasets for classification and segmentation.
The following section provides an overview of the included tables and highlights their aspects that may be of particular interest to the reader.
\begin{itemize}
    \item \textbf{Tab.~\ref{tab_ap:4t}} extends the evaluation of classifiers trained from scratch (MetaFormer with architecture signature $[T$, $T$, $T$, $T]$ and ResNet18) by including accuracy and F1-score.
    This table provides additional insights into the results discussed in Sec.~\ref{sec:classification_4t}.
    \item \textbf{Tab.~\ref{tab_ap:2p2t_pretrained}} extends the evaluation of classifiers utilizing ImageNet-pretrained weights (MetaFormer with architecture signature $[P, P, T, T]$ and ResNet18) to include accuracy and F1-score.  
    It further complements the discussion presented in Sec.~\ref{sec:classification_2p2t}.
    \item \textbf{Tab.~\ref{tab_ap:2p2t_scratch}} reports the \ac{auc} and F1-score for the MetaFormer $[P, P, T, T]$ variant trained from scratch.  
    Based on these results and those from Tab.~\ref{tab:classification} (meta-row $[P, P, T, T]$), the percentage uplift values when compared to pretrained weights are calculated as shown in Fig.~\ref{fig:2p2t_uplift}.
    \item \textbf{Tab.~\ref{tab_ap:classification_ranking}} presents the absolute and relative ranking scores of different token mixers across classification datasets.  
    This information may be particularly useful for readers seeking the best-performing token mixers for their specific domains.
    \item Similar to Tab.~\ref{tab_ap:classification_ranking}, \textbf{Tab.~\ref{tab_ap:seg_ranking}} reports the absolute and relative ranking scores of token mixers across segmentation datasets.
\end{itemize}

\section{MetaFormer Architecture S-Variants}\label{sec_ap:s_variants}
We include the MetaFormer variants S24 and S36 across selected datasets and token mixers to gather a first impression whether our findings depend on model scale.
For classification, we focus on DermaMNIST, identified as the most challenging with the S12 variant, using grouped convolution with kernel size $K=3$ for local inductive bias and random token mixing for global context, which performed second-best in the S12 setting.
Given DermaMNIST’s limited size ($7\,mathrm{K}$ training images) and the increased parameter count of larger variants, we also evaluate on PathMNIST ($90\,\mathrm{K}$ images), using the best-performing mixers: convolution and pooling with $K=7$.
As shown in Tab.~\ref{tab_ap:s_variants}, the resulting \acf{auc} consistently decreases with larger model variants, suggesting that increased depth and capacity are not beneficial in this setting.\par
For segmentation, we select AbdomenAtlas as the largest dataset and to include a 3D task, using pooling and grouped convolution (both with $K=3$) to remain within memory constraints. While no clear trend is observed for pooling, grouped convolution benefits from increased model scale.
This suggests that AbdomenAtlas dataset, with its $4\,675$ 3D training images, supports the optimization of parameter-heavy models, as already reflected by the superior performance of standard convolution over grouped convolution in the S12 setting (cf. Tab.~\ref{tab:3d_task}).

\section{ImageWoof}\label{sec_ap:imagewoof}
To measure the factor of a potential domain shift between the pretrained ImageNet weights and the medical data, we also include a small subset of the ImageNet dataset, since a full-size ImageNet training would have been unfeasible considering the number of experiments and available hardware.
We settle on ImageWoof \citep{HowardImagewoof2019}: A non-trivial dataset of ten different dog breeds, coming with a predefined split of $9\,035/3\,939$ train/test images.
For training from scratch ($[T,T,T,T]$), we find token mixers with additional learnable parameters decrease the performance, instead identity and pooling yield the best \ac{auc} (Tab.~\ref{tab_ap:4t}).
For transfer learning ($[P, P, T, T]$ experiemnts in Tab.~\ref{tab_ap:2p2t_scratch} and Tab.~\ref{tab_ap:2p2t_pretrained}), ImageWoof provides a proof of concept, that attention weights of global attention can be effectively reused in local-attention (warm start).
As expected, this effect becomes more pronounced as the kernel size of local attention increases, making the two operations more similar.
Beyond the benefit of reusing attention weights ($+10.54\,\%$, $+13.95\,\%$, $+15.53\,\%$), employing pretrained weights for the channel-\acp{mlp} alone also yields a substantial improvement ($+9.94\,\%$, $+11.08\,\%$, $+12.13\,\%$) for local attention trained from scratch.
The performance gain is as strong for global attention with ($+11.37\,\%$) and without warm start ($+8.29\,\%$).
We observe performance gains from pre-training across all other token mixers even if the introduced domain shift diminishes this effect.
Notably, the magnitude of improvement correlates with the similarity to the original global self-attention token mixer: for example, parameterized convolutions with the largest kernels possibly benefit more than non-parameterized pooling or no token mixing (identity).
The average improvement of $+7.06\,\%$ indicates that transferring pretrained weights to different token mixers is very effective when staying within the natural image domain.

\begin{figure}[b]
    \centering
    \includegraphics[width=0.25\textwidth]{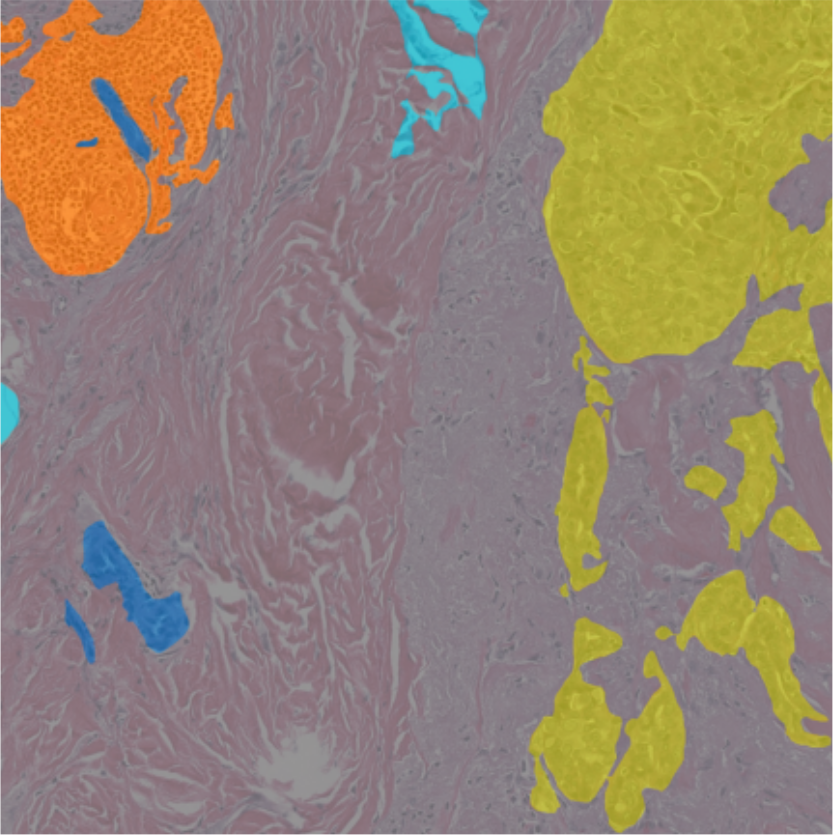}
    \caption{Ground truth segmentation mask for the test case of the TIGER dataset, as shown in Fig.~\ref{fig_ap:qualitative_segmentation_results}.}
    \label{fig:tiger_seg_ground_truth}
\end{figure}

\begin{sidewaystable*}
    \centering
    \caption{Scaling of the MetaFormer S-variants trained from scratch ($[T,T,T,T]$ for a selection of token mixers (trainable vs non-trainable, local vs global). For classification, we chose the most challenging task (Derma) and the largest one (Path with $90\,k$ training images). For segmentation, we choose the Abdomen Atlas due to its size and to include a 3D task.}
    \label{tab_ap:s_variants}
    \begin{tabular}{llc|ccc|c}
    \toprule
     Dataset & Token Mixer & Kernel & S12 & S24 & S36 & \#P[M] \\
    \midrule
    \multicolumn{7}{c}{2D Classification (Acc, \acs{auc}, F1)}\\
    \midrule
    \multirow[c]{2}{*}{Derma} & grouped conv & 3 & 0.7138, 0.8867, 0.7029 & 0.7048, 0.8745, 0.7136 & 0.7064, 0.8742, 0.7170&11.4/20.9/30.5\\
    & random & - & 0.7070, 0.9024, 0.7230 & 0.6766, 0.8673, 0.7059 & 0.6782, 0.8456, 0.6914 & 11.4/20.9/30.4\\
    \midrule
    \multirow[c]{2}{*}{Path} & pooling & 7 & 0.9030, 0.9829, 0.8986 & 0.9085, 0.982, 0.9027 & 0.8935, 0.9823, 0.8934 & 11.4/20.9/30.4\\
    & conv & 7 & 0.9115, 0.9887, 0.9112 & 0.8968, 0.9851, 0.8970 & 0.9136, 0.9877, 0.9151 & 69.2/136/203\\
    \midrule
    \multicolumn{7}{c}{3D Segmentation (\ac{dsc})}\\
    \midrule
    \multirow[c]{2}{*}{Abdomen} & pooling & 3 & 0.9405 & 0.9410 & 0.9386 & 15.8/25.3/34.7\\
    & grouped conv & 3 & 0.9379 & 0.9454 & 0.9431 & 15.9/25.4/35.0\\
    \bottomrule
    \end{tabular}
\end{sidewaystable*}

\begin{sidewaystable*}[ht]
    \centering
    \caption{Performance for different token mixers $T$ across datasets. Each stage of the MetaFormer utilizes the given token mixer and was trained from scratch. Due to training instabilities within the implementation of local attention, we had to adapt the training routine (cf. Sec.~\ref{sec:training_instability}). Please refer to markers (*,+,$\dagger$) in Tab.~\ref{tab:classification} metarow $[T,T,T,T]$ to identify those results.}
    \begin{tabular}{lc|cc|cc|cc|cc|cc}
    \toprule
     &  & \multicolumn{2}{c}{ImageWoof} & \multicolumn{2}{c}{PathMNIST} & \multicolumn{2}{c}{DermaMNIST} & \multicolumn{2}{c}{PneumoniaMNIST} & \multicolumn{2}{c}{OrganSMNIST} \\
    Token Mixer $T$ & Kernel $K$ & Acc & F1 & Acc & F1 & Acc & F1 & Acc & F1 & Acc & F1 \\
    \midrule
    \multirow[c]{3}{*}{pooling} & 3 & 0.7908 & 0.7907 & 0.8817 & 0.8767 & 0.6828 & 0.6868 & 0.9632 & 0.9656 & 0.7659 & 0.7689 \\
     & 5 & 0.7941 & 0.7965 & 0.8832 & 0.8773 & 0.6933 & 0.6855 & 0.9393 & 0.9446 & 0.7494 & 0.7576 \\
     & 7 & 0.8044 & 0.8059 & 0.9030 & 0.8986 & 0.6811 & 0.6991 & 0.9526 & 0.9537 & 0.7649 & 0.7706 \\
     \midrule
    \multirow[c]{3}{*}{conv} & 3 & 0.7869 & 0.7891 & 0.8862 & 0.8814 & 0.7083 & 0.7123 & 0.9064 & 0.9201 & 0.7721 & 0.7765 \\
     & 5 & 0.7540 & 0.7567 & 0.9045 & 0.9035 & 0.7012 & 0.7013 & 0.9419 & 0.9480 & 0.7695 & 0.7743 \\
     & 7 & 0.7078 & 0.7090 & 0.9115 & 0.9112 & 0.6883 & 0.6872 & 0.9500 & 0.9535 & 0.7872 & 0.7901 \\
     \midrule
    \multirow[c]{3}{*}{grouped conv} & 3 & 0.7915 & 0.7937 & 0.8888 & 0.8840 & 0.7138 & 0.7029 & 0.9474 & 0.9532 & 0.7837 & 0.7863 \\
     & 5 & 0.7694 & 0.7713 & 0.9068 & 0.9065 & 0.7091 & 0.7062 & 0.9534 & 0.9569 & 0.7961 & 0.8001 \\
     & 7 & 0.7411 & 0.7424 & 0.8927 & 0.8884 & 0.6882 & 0.7121 & 0.9521 & 0.9552 & 0.7850 & 0.7871 \\
     \midrule
    \multirow[c]{3}{*}{local attn} & 3 & 0.7357 & 0.7371 & 0.8580 & 0.8476 & 0.7255 & 0.7263 & 0.9521 & 0.9521 & 0.7852 & 0.7868 \\
     & 5 & 0.7203 & 0.7206 & 0.8868 & 0.8781 & 0.6999 & 0.7199 & 0.9333 & 0.9408 & 0.7830 & 0.7875 \\
     & 7 & 0.6914 & 0.6908 & 0.8794 & 0.8728 & 0.6920 & 0.7081 & 0.9346 & 0.9395 & 0.7826 & 0.7866 \\
     \midrule
    global attn & - & 0.5439 & 0.5408 & 0.8778 & 0.8647 & 0.6046 & 0.5384 & 0.8688 & 0.8810 & 0.7357 & 0.7410 \\
    \midrule
    random & - & 0.7900 & 0.7911 & 0.9072 & 0.9039 & 0.707 & 0.723 & 0.9594 & 0.9637 & 0.7709 & 0.7731 \\
    \midrule
    identity & 1 & 0.7695 & 0.7698 & 0.8913 & 0.8822 & 0.6722 & 0.6761 & 0.9432 & 0.9466 & 0.7687 & 0.7720 \\
    \midrule
    \multirow[c]{3}{*}{ResNet18} & 3 & 0.7454 & 0.7478 & 0.8392 & 0.8260 & 0.7332 & 0.7388 & 0.8885 & 0.9048 & 0.7866 & 0.7904 \\
     & 5 & 0.7257 & 0.7271 & 0.8732 & 0.8615 & 0.7002 & 0.7331 & 0.9308 & 0.9404 & 0.7836 & 0.7907 \\
     & 7 & 0.6909 & 0.6928 & 0.8619 & 0.8496 & 0.6912 & 0.7015 & 0.9124 & 0.9241 & 0.7763 & 0.7832 \\
    \bottomrule
    \end{tabular}
    \label{tab_ap:4t}
\end{sidewaystable*}

\begin{sidewaystable*}[ht]
    \centering
    \caption{Performance for different token mixers $T$ across datasets. The MetaFormer utilizes ImageNet-pretrained weights for the channel-MLPs and only exchanges token mixers in the last two stages, keeping the original pooling at the first two stages ($[P,P,T,T]$).}
    \begin{tabular}{lc|cc|cc|cc|cc|cc}
    \toprule
     &  & \multicolumn{2}{c}{ImageWoof} & \multicolumn{2}{c}{PathMNIST} & \multicolumn{2}{c}{DermaMNIST} & \multicolumn{2}{c}{PneumoniaMNIST} & \multicolumn{2}{c}{OrganSMNIST} \\
    Token Mixer $T$ & Kernel $K$ & Acc & F1 & Acc & F1 & Acc & F1 & Acc & F1 & Acc & F1 \\
    \midrule
    \multirow[c]{3}{*}{pooling $P$} & 3 & 0.8501 & 0.8531 & 0.8643 & 0.8539 & 0.7585 & 0.7737 & 0.9321 & 0.9421 & 0.7620 & 0.7657 \\
     & 5 & 0.8610 & 0.8637 & 0.8726 & 0.8647 & 0.7232 & 0.7625 & 0.9521 & 0.9584 & 0.7632 & 0.7648 \\
     & 7 & 0.8612 & 0.8633 & 0.8914 & 0.8869 & 0.7789 & 0.8058 & 0.9423 & 0.9496 & 0.7725 & 0.7750 \\
     \midrule
    \multirow[c]{3}{*}{conv} & 3 & 0.8494 & 0.8524 & 0.8768 & 0.8687 & 0.7523 & 0.7624 & 0.9444 & 0.9513 & 0.7889 & 0.7908 \\
     & 5 & 0.8361 & 0.8383 & 0.8818 & 0.8705 & 0.7642 & 0.7894 & 0.9372 & 0.9459 & 0.7963 & 0.7967 \\
     & 7 & 0.8259 & 0.8256 & 0.8684 & 0.8570 & 0.7863 & 0.7818 & 0.9517 & 0.9568 & 0.7740 & 0.7794 \\
     \midrule
    \multirow[c]{3}{*}{grouped conv} & 3 & 0.8491 & 0.8521 & 0.8601 & 0.8475 & 0.7740 & 0.7936 & 0.9487 & 0.9549 & 0.7818 & 0.7831 \\
     & 5 & 0.8308 & 0.8338 & 0.8977 & 0.8949 & 0.7594 & 0.7928 & 0.9598 & 0.9654 & 0.7916 & 0.7973 \\
     & 7 & 0.8375 & 0.8400 & 0.8967 & 0.8919 & 0.7605 & 0.7916 & 0.9214 & 0.9330 & 0.7910 & 0.7965 \\
     \midrule
    \multirow[c]{3}{*}{local attn} & 3 & 0.8735 & 0.8753 & 0.8856 & 0.8775 & 0.7737 & 0.7923 & 0.9321 & 0.9421 & 0.7747 & 0.7790 \\
     & 5 & 0.8614 & 0.8619 & 0.8852 & 0.8786 & 0.7631 & 0.7613 & 0.9513 & 0.9552 & 0.7728 & 0.7790 \\
     & 7 & 0.8679 & 0.8697 & 0.8868 & 0.8784 & 0.7577 & 0.7738 & 0.9513 & 0.9583 & 0.7617 & 0.7667 \\
     \midrule
    \multirow[c]{3}{*}{local attn (warm start)} & 3 & 0.8742 & 0.8759 & 0.8918 & 0.8876 & 0.7899 & 0.7995 & 0.9607 & 0.9654 & 0.7769 & 0.7814 \\
     & 5 & 0.8841 & 0.8875 & 0.8884 & 0.8829 & 0.7879 & 0.8035 & 0.9632 & 0.9688 & 0.7921 & 0.7956 \\
     & 7 & 0.8834 & 0.8841 & 0.8991 & 0.8931 & 0.7778 & 0.7895 & 0.9436 & 0.9482 & 0.7862 & 0.7918 \\
     \midrule
    global attn & - & 0.8641 & 0.8662 & 0.8995 & 0.8962 & 0.7451 & 0.7714 & 0.9286 & 0.9386 & 0.7667 & 0.7732 \\
    global attn (warm start) & - & 0.8908 & 0.8939 & 0.9011 & 0.8990 & 0.7672 & 0.7821 & 0.9444 & 0.9513 & 0.7792 & 0.7850 \\
    \midrule
    random & - & 0.8553 & 0.8571 & 0.8994 & 0.8969 & 0.7713 & 0.781 & 0.9329 & 0.9423 & 0.7687 & 0.7732\\
    \midrule
    identity & 1 & 0.8597 & 0.8604 & 0.7985 & 0.7792 & 0.8056 & 0.8077 & 0.9291 & 0.9402 & 0.7658 & 0.7695 \\
    \midrule
    ResNet18 & 3 & 0.8968 & 0.8989 & 0.8976 & 0.8900 & 0.8071 & 0.8227 & 0.8974 & 0.9139 & 0.7920 & 0.7955 \\
    \bottomrule
    \end{tabular}
    \label{tab_ap:2p2t_pretrained}
\end{sidewaystable*}

\begin{sidewaystable*}[ht]
    \centering
    \caption{Performance (\textbf{\acs{auc}} and F1) for different token mixers $T$ across datasets. The MetaFormer utilizes pooling at the first two stages and the given token mixer at the last two stages ($[P,P,T,T]$). These results provide the base for the calculation of  the percental improvement shown in Fig.~\ref{fig:2p2t_uplift}.}
    \begin{tabular}{lc|cc|cc|cc|cc|cc}
    \toprule
     &  & \multicolumn{2}{c}{ImageWoof} & \multicolumn{2}{c}{PathMNIST} & \multicolumn{2}{c}{DermaMNIST} & \multicolumn{2}{c}{PneumoniaMNIST} & \multicolumn{2}{c}{OrganSMNIST} \\
    Token Mixer $T$ & Kernel $K$ & AUC & F1 & AUC & F1 & AUC & F1 & AUC & F1 & AUC & F1 \\
    \midrule
    \multirow[c]{3}{*}{pooling} & 3 & 0.9495 & 0.7907 & 0.9758 & 0.8767 & 0.8883 & 0.6868 & 0.9921 & 0.9656 & 0.9617 & 0.7689 \\
     & 5 & 0.9517 & 0.7965 & 0.9799 & 0.8773 & 0.8911 & 0.6855 & 0.9832 & 0.9446 & 0.9375 & 0.7576 \\
     & 7 & 0.9539 & 0.8059 & 0.9829 & 0.8986 & 0.8795 & 0.6991 & 0.9814 & 0.9537 & 0.9420 & 0.7706 \\
     \midrule
    \multirow[c]{3}{*}{conv} & 3 & 0.9466 & 0.7657 & 0.9697 & 0.8551 & 0.9035 & 0.7141 & 0.9755 & 0.9425 & 0.9594 & 0.7804 \\
     & 5 & 0.9191 & 0.7243 & 0.9785 & 0.8735 & 0.8456 & 0.6869 & 0.9841 & 0.9386 & 0.9611 & 0.7921 \\
     & 7 & 0.9036 & 0.6752 & 0.9862 & 0.9145 & 0.8526 & 0.6774 & 0.9864 & 0.9496 & 0.9671 & 0.7789 \\
     \midrule
    \multirow[c]{3}{*}{grouped conv} & 3 & 0.9357 & 0.7616 & 0.9708 & 0.8538 & 0.8865 & 0.7174 & 0.9894 & 0.9654 & 0.9468 & 0.7922 \\
     & 5 & 0.9208 & 0.7166 & 0.9826 & 0.9066 & 0.8793 & 0.6845 & 0.9895 & 0.9569 & 0.9474 & 0.7798 \\
     & 7 & 0.9089 & 0.6836 & 0.9846 & 0.8991 & 0.8607 & 0.7012 & 0.9857 & 0.9353 & 0.9506 & 0.7918 \\
     \midrule
    \multirow[c]{3}{*}{local attn} & 3 & 0.8796 & 0.5225 & 0.9771 & 0.8529 & 0.9168 & 0.5749 & 0.9716 & 0.9083 & 0.9702 & 0.7562 \\
     & 5 & 0.8691 & 0.4881 & 0.9690 & 0.8074 & 0.9179 & 0.5452 & 0.9734 & 0.9171 & 0.9721 & 0.7598 \\
     & 7 & 0.8620 & 0.4783 & 0.9733 & 0.8096 & 0.9230 & 0.5553 & 0.9665 & 0.9082 & 0.9701 & 0.7447 \\
     \midrule
    global attn & - & 0.8907 & 0.5486 & 0.9823 & 0.8470 & 0.9328 & 0.5944 & 0.9638 & 0.8582 & 0.9686 & 0.7646 \\
    \midrule
    random & - & 0.9457 & 0.7708 & 0.9822 & 0.9026 & 0.9115 & 0.7024 & 0.9863 & 0.9389 & 0.952 & 0.7773\\
    \midrule
    identity & 1 & 0.9659 & 0.7852 & 0.9560 & 0.7465 & 0.9474 & 0.7106 & 0.9892 & 0.9533 & 0.9637 & 0.7750 \\
    \bottomrule
    \end{tabular}
    \label{tab_ap:2p2t_scratch}
\end{sidewaystable*}

\begin{table*}[ht]
    \centering
    \caption{Detailed results by our ranking schema (cf. Sec.~\ref{sec:ranking}) for each token mixer across datasets. Each entry shows a tuple: the first element is the absolute number of pairwise wins against other token mixers, and the second is the normalized rank in $[0.1, 1]$. Global ranks are obtained as the geometric mean of normalized ranks across datasets.}
    \begin{tabular}{r|lc|cccc|c}
    \toprule
    & \multicolumn{2}{r|}{Dataset} & Path & Derma & Pneumonia & OrganS & \multirow[c]{2}{5em}{\centering geometric mean}\\
    \cmidrule{2-3}
    & Token Mixer $T$ & Kernel $K$ & & & & & \\ 
    \midrule
    \multirow{14}{*}{\rotatebox[origin=c]{90}{$[T, T, T, T]$ from scratch}}&\multirow[c]{3}{*}{pooling} & 3 & (2, 0.32) & (2, 0.74) & (5, 0.9) & (12, 0.9) & 0.666 \\
     && 5 & (4, 0.55) & (2, 0.74) & (1, 0.42) & (1, 0.2) & 0.429 \\
     && 7 & (8, 0.81) & (1, 0.49) & (1, 0.42) & (2, 0.36) & 0.493 \\
     \cmidrule{2-8}
    &\multirow[c]{3}{*}{conv} & 3 & (10, 0.87) & (0, 0.23) & (1, 0.42) & (9, 0.77) & 0.505 \\
     && 5 & \textbf{(13, 0.97)} & (2, 0.74) & (1, 0.42) & (6, 0.61) & 0.657 \\
     && 7 & \textbf{(13, 0.97)} & (0, 0.23) & \textbf{(7, 1.0)} & (7, 0.68) & 0.622 \\
     \cmidrule{2-8}
    &\multirow[c]{3}{*}{grouped conv} & 3 & (4, 0.55) & (2, 0.74) & (5, 0.9) & (9, 0.77) & \textbf{0.731} \\
     && 5 & (6, 0.74) & (1, 0.49) & (1, 0.42) & (2, 0.36) & 0.483 \\
     && 7 & (4, 0.55) & (0, 0.23) & (3, 0.77) & (4, 0.49) & 0.466 \\
     \cmidrule{2-8}
    &\multirow[c]{3}{*}{local attn} & 3 & (0, 0.13) & (2, 0.74) & (1, 0.42) & (2, 0.36) & 0.349 \\
     && 5 & (0, 0.13) & (0, 0.23) & (1, 0.42) & (0, 0.1) & 0.189 \\
     && 7 & (2, 0.32) & (0, 0.23) & (1, 0.42) & (1, 0.2) & 0.28 \\
     \cmidrule{2-8}
    & global attn & - & (4, 0.55) & \textbf{(13, 1.0)} & (0, 0.1) & \textbf{(14, 1.0)} & 0.484 \\
    \cmidrule{2-8}
    & random & - & (4, 0.55) & (6, 0.94) & (1, 0.42) & (5, 0.55) & 0.588 \\
    \cmidrule{2-8}
    & identity & 1 & (1, 0.23) & (1, 0.49) & (3, 0.77) & (12, 0.9) & 0.528 \\
    \midrule
    \multirow{18}{*}{\rotatebox[origin=c]{90}{$[P, P, T, T]$ pretrained channel-MLPs}}&\multirow[c]{3}{*}{pooling $P$} & 3 & (1, 0.17) & (0, 0.5) & (0, 0.37) & (7, 0.55) & 0.367 \\
     && 5 & (3, 0.35) & (0, 0.5) & (0, 0.37) & (15, 0.95) & 0.5 \\
     && 7 & (8, 0.75) & (0, 0.5) & (0, 0.37) & (2, 0.38) & 0.479 \\
     \cmidrule{2-8}
    &\multirow[c]{3}{*}{conv} & 3 & (1, 0.17) & (0, 0.5) & (0, 0.37) & (7, 0.55) & 0.367 \\
     && 5 & (4, 0.4) & (0, 0.5) & (0, 0.37) & (12, 0.9) & 0.51 \\
     && 7 & (7, 0.55) & (0, 0.5) & (0, 0.37) & (8, 0.73) & 0.523 \\
     \cmidrule{2-8}
    &\multirow[c]{3}{*}{grouped conv} & 3 & (2, 0.27) & (0, 0.5) & \textbf{(5, 0.95)} & (1, 0.22) & 0.414 \\
     && 5 & (2, 0.27) & (0, 0.5) & (4, 0.83) & (7, 0.55) & 0.5 \\
     && 7 & \textbf{(15, 0.9)} & (0, 0.5) & (0, 0.37) & (1, 0.22) & 0.441 \\
     \cmidrule{2-8}
    &\multirow[c]{3}{*}{local attn} & 3 & (8, 0.75) & (0, 0.5) & (0, 0.37) & (2, 0.38) & 0.479 \\
     && 5 & (7, 0.55) & (0, 0.5) & (8, 1.0) & (8, 0.73) & \textbf{0.668} \\
     && 7 & (7, 0.55) & (0, 0.5) & (0, 0.37) & (8, 0.73) & 0.523 \\
     \cmidrule{2-8}
    &\multirow[c]{3}{6em}{local attn (warm start)} & 3 & (8, 0.75) & (0, 0.5) & (0, 0.37) & (9, 0.85) & 0.588 \\
     && 5 & (7, 0.55) & (0, 0.5) & (4, 0.83) & (1, 0.22) & 0.475 \\
     && 7 & \textbf{(15, 0.9)} & (0, 0.5) & (0, 0.37) & \textbf{(16, 1.0)} & 0.641 \\
     \cmidrule{2-8}
    &global attn & - & \textbf{(15, 0.9)} & (0, 0.5) & (1, 0.7) & (5, 0.45) & 0.614 \\
    &(warm start) & - & (18, 1.0) & (0, 0.5) & (4, 0.83) & (0, 0.1) & 0.451 \\
    \cmidrule{2-8}
    &random & - & (7, 0.55) & (17, 0.95) & (0, 0.37) & (1, 0.22) & 0.458 \\
    \cmidrule{2-8}
    &identity & 1 & (0, 0.1) & \textbf{(18, 1.0)} & (4, 0.83) & (8, 0.73) & 0.495 \\
    \bottomrule
\end{tabular}
    \label{tab_ap:classification_ranking}
\end{table*}

\begin{table*}[ht]
    \centering
    \caption{Detailed results by our ranking schema (cf. Sec.~\ref{sec:ranking}) for each token mixer across datasets. Each entry shows a tuple: the first element is the absolute number of pairwise wins against other token mixers, and the second is the normalized rank in $[0.1, 1]$. Global ranks are obtained as the geometric mean of normalized ranks across datasets.}
    \begin{tabular}{c|lc|ccc|c}
    \toprule
     &\multicolumn{2}{r|}{Dataset}  & JSRT & GRAZ & TIGER@768² & \\
     \cmidrule{2-3}
     &TokenMixer $T$ & Kernel $K$ & & & & geometric mean \\
     \midrule
    \multirow{18}{*}{\rotatebox[origin=c]{90}{$[T, T, T, T]$ from scratch}}&\multirow[c]{4}{*}{pooling} & 3 & (4, 0.52) & (3, 0.49) & (7, 0.92) & 0.614 \\
     && 5 & (3, 0.42) & (3, 0.49) & (3, 0.58) & 0.49 \\
     && 7 & (1, 0.26) & (3, 0.49) & (3, 0.58) & 0.418 \\
     && 9 &  &  & (7, 0.92) & \\
     \cmidrule{2-7}
    &\multirow[c]{4}{*}{conv} & 3 & (8, 0.68) & (6, 0.74) & \textbf{(11, 1.0)} & 0.796 \\
     && 5 & \textbf{(11, 1.0)} & (7, 0.81) & (2, 0.44) & 0.71 \\
     && 7 & (10, 0.87) & \textbf{(11, 1.0)} & (2, 0.44) & 0.729 \\
     && 9 &  & & (4, 0.71) & \\
     \cmidrule{2-7}
    &\multirow[c]{4}{*}{grouped conv} & 3 & (10, 0.87) & (3, 0.49) & (3, 0.58) & 0.625 \\
     && 5 & (9, 0.74) & (10, 0.94) & (6, 0.84) & 0.836 \\
     && 7 & (10, 0.87) & (9, 0.87) & (5, 0.79) & \textbf{0.843} \\
     && 9 &  &  & (4, 0.71) & \\
     \cmidrule{2-7}
    &\multirow[c]{4}{*}{local attn} & 3 & (0, 0.1) & (3, 0.49) & (0, 0.23) & 0.224 \\
     && 5 & (4, 0.52) & (3, 0.49) & (0, 0.23) & 0.388 \\
     && 7 & (1, 0.26) & (3, 0.49) & (0, 0.23) & 0.309 \\
     && 9 &  &  & (0, 0.23) & \\
     \cmidrule{2-7}
    &global attn & - & (1, 0.26) & (0, 0.16) & & 0.207 \\
    \cmidrule{2-7}
    &random & - & (5, 0.61) & (0, 0.16) & (0, 0.23) & 0.286 \\
    \cmidrule{2-7}
    &identity & 1 & (1, 0.26) & (0, 0.16) & (0, 0.23) & 0.215 \\
     \bottomrule
    \end{tabular}
    \label{tab_ap:seg_ranking}
\end{table*}

\begin{figure*}[h]
    \centering
    \begin{subfigure}{\textwidth}
        \centering
        \caption*{\textbf{JSRT}}
        \begin{subfigure}[t]{.24\linewidth}
            \includegraphics[width=\textwidth]{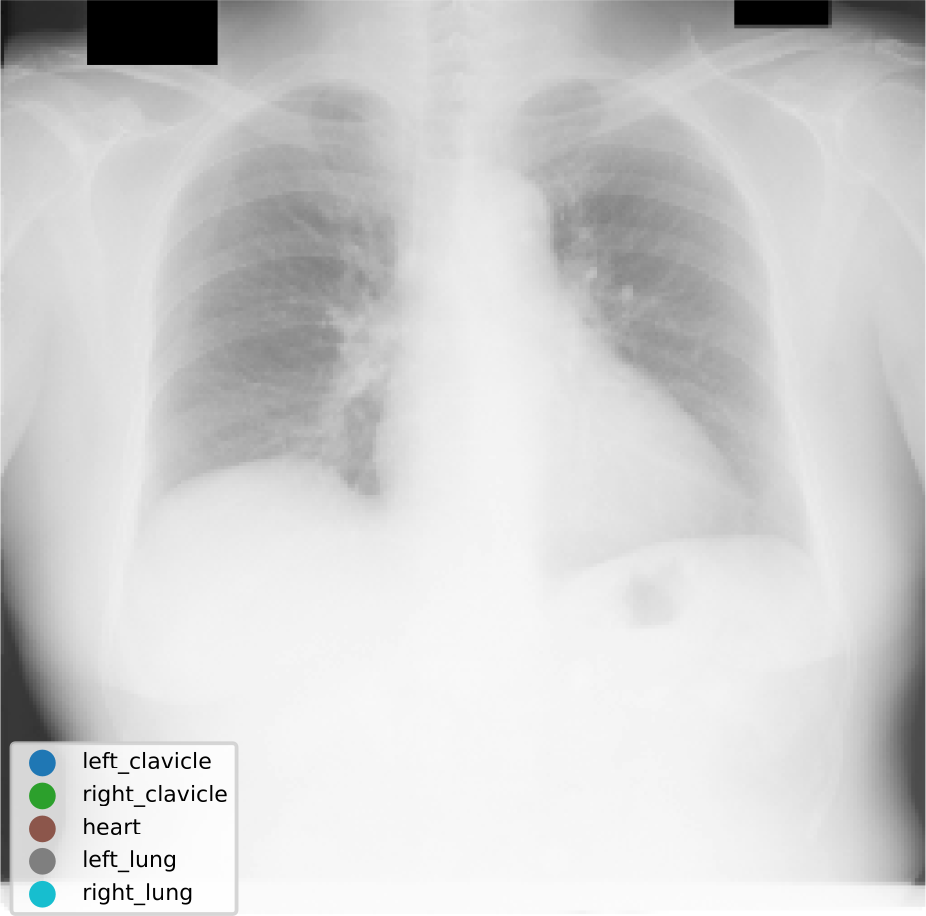}
        \end{subfigure}\hfill
        \begin{subfigure}[t]{.24\textwidth} 
            \includegraphics[width=\textwidth]{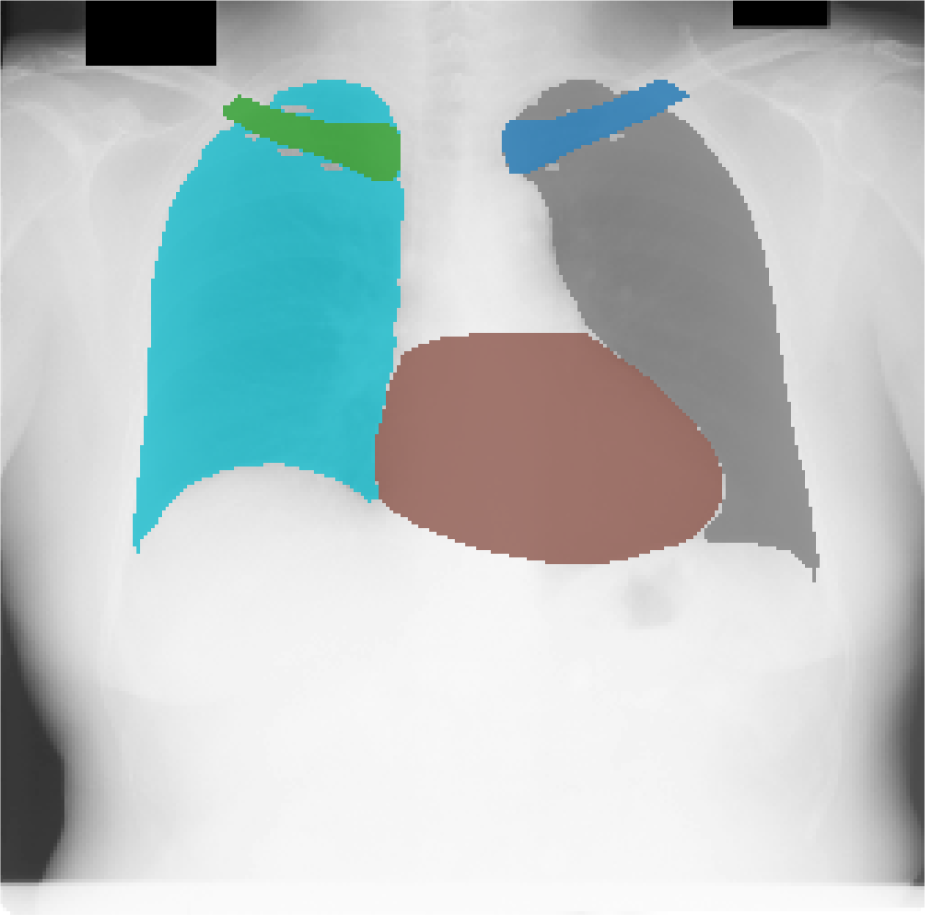}
            \caption{\begin{tabular}{l}
                 pooling (3)  \\
                 $0.9482\pm0.0207$
            \end{tabular}}
        \end{subfigure}\hfill
        \begin{subfigure}[t]{.24\textwidth}
            \includegraphics[width=\textwidth]{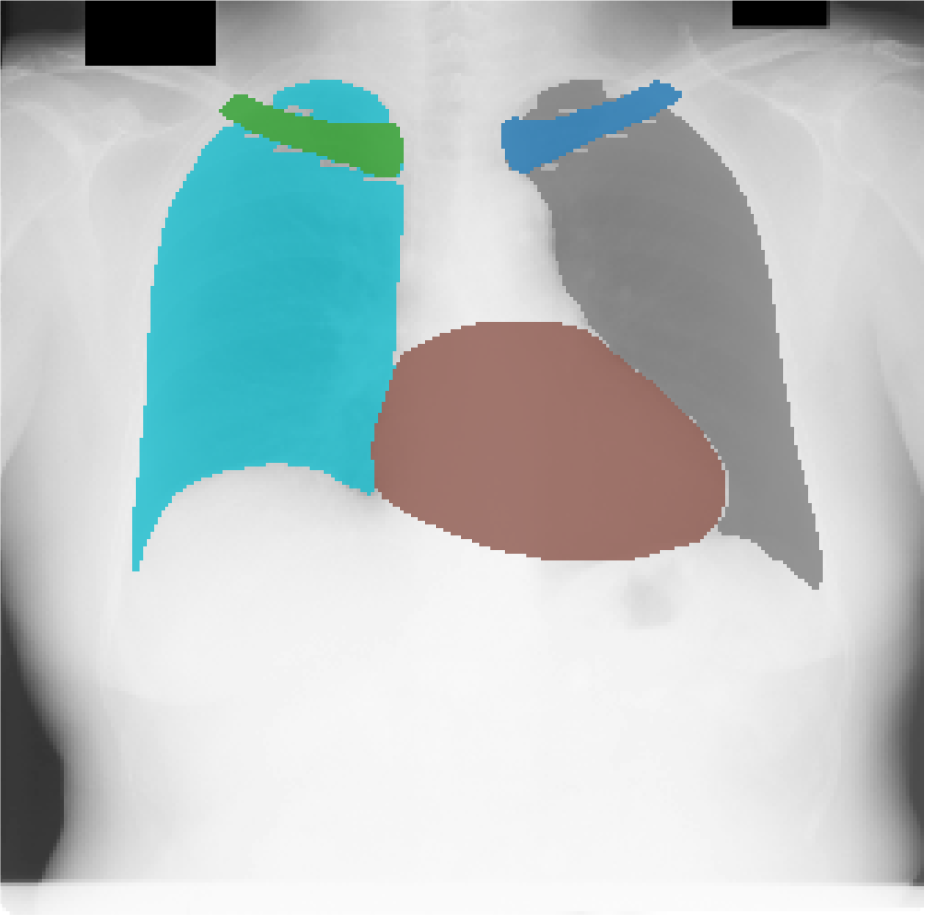}
            \caption{\begin{tabular}{l}
                 grouped conv (7)  \\
                 $0.9516\pm0.017$
            \end{tabular}}
        \end{subfigure}\hfill
        \begin{subfigure}[t]{.24\textwidth}
            \includegraphics[width=\textwidth]{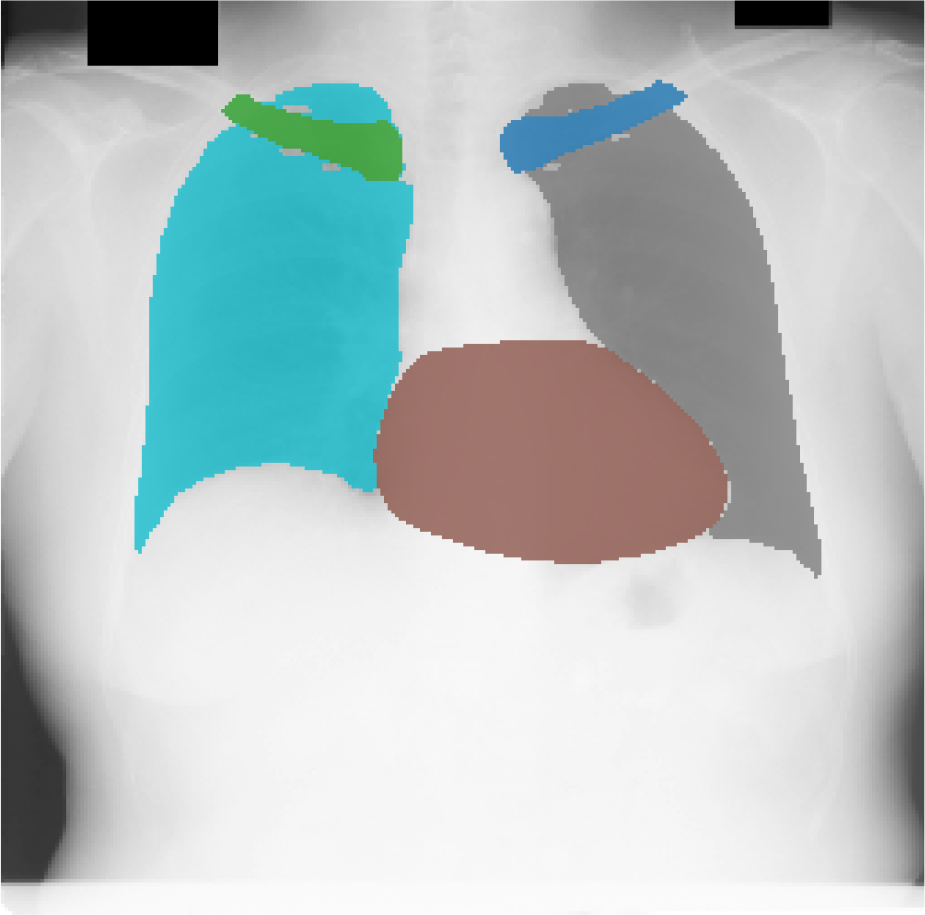}
            \caption{\begin{tabular}{l}
                 random  \\
                 $0.9505\pm0.020$
            \end{tabular}}
        \end{subfigure}
    \end{subfigure}
    \begin{subfigure}{\textwidth}
        \centering
        \caption*{\textbf{GRAZ}}
        \begin{subfigure}[t]{.24\linewidth}
            \includegraphics[width=\textwidth]{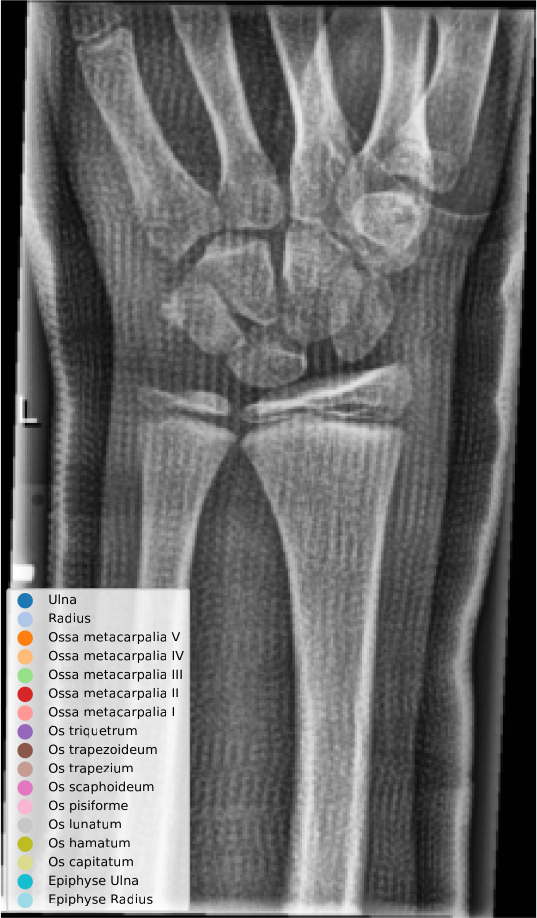}
        \end{subfigure}\hfill
        \begin{subfigure}[t]{.24\textwidth} 
            \includegraphics[width=\textwidth]{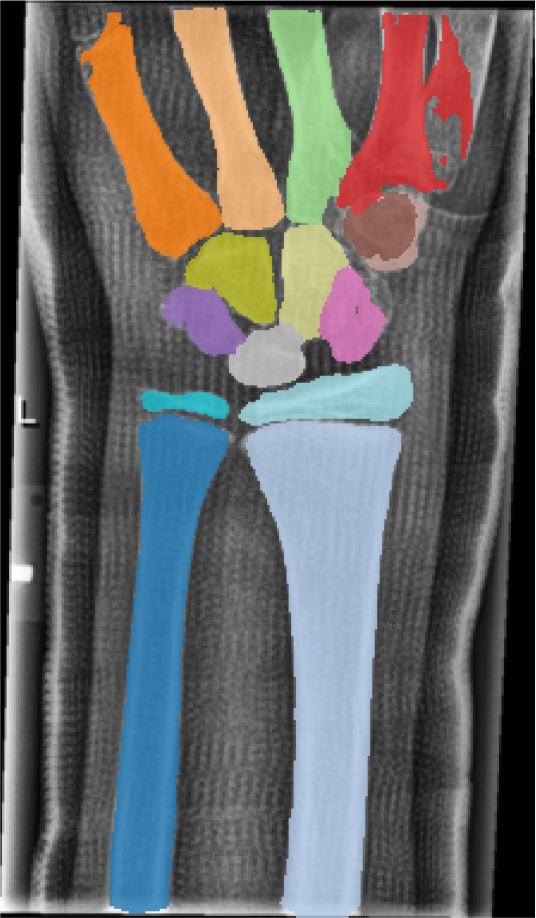}
            \caption{\begin{tabular}{l}
                 pooling (3)  \\
                 $0.8460\pm0.2275$
            \end{tabular}}
        \end{subfigure}\hfill
        \begin{subfigure}[t]{.24\textwidth}
            \includegraphics[width=\textwidth]{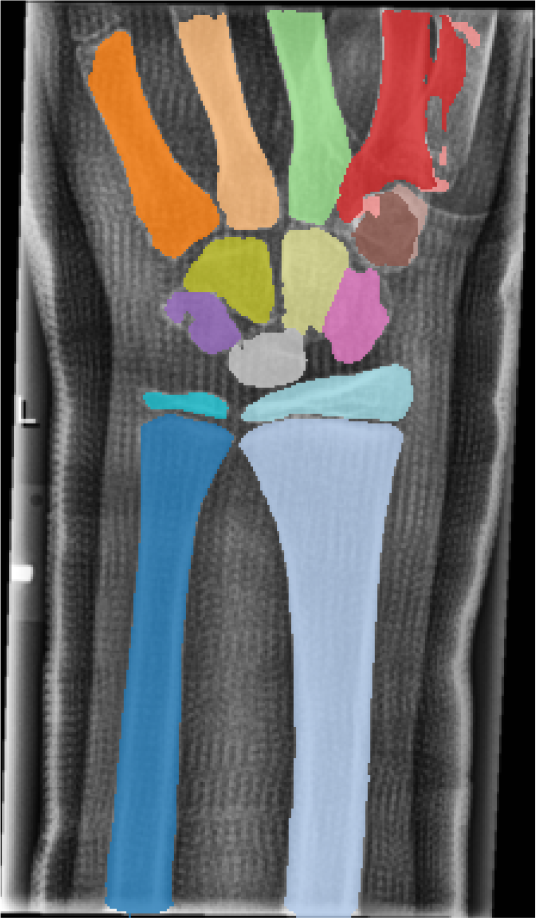}
            \caption{\begin{tabular}{l}
                 grouped conv (7)  \\
                 $0.8401\pm0.1824$
            \end{tabular}}
        \end{subfigure}\hfill
        \begin{subfigure}[t]{.24\textwidth}
            \includegraphics[width=\textwidth]{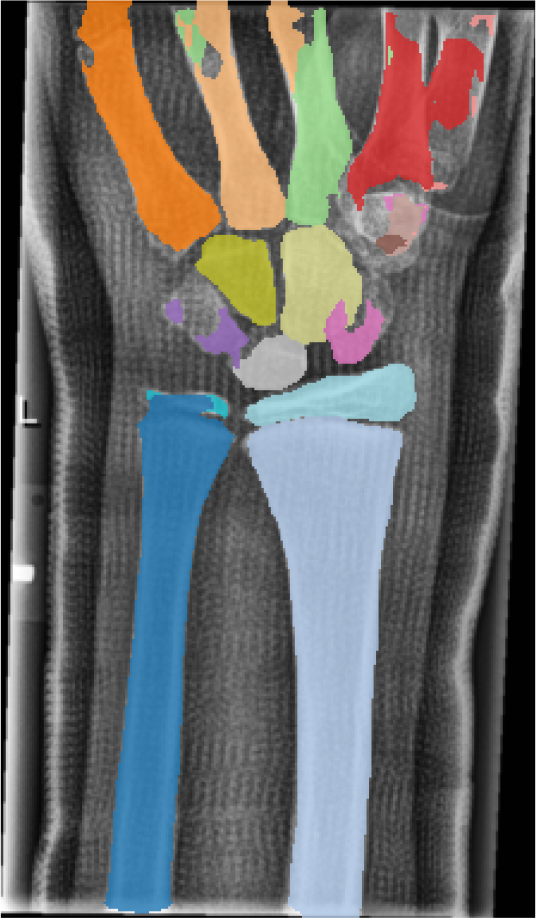}
            \caption{\begin{tabular}{l}
                 random  \\
                 $0.6980\pm0.2651$
            \end{tabular}}
        \end{subfigure}
    \end{subfigure}
    \begin{subfigure}{\textwidth}
        \centering
        \caption*{\textbf{TIGER}}
        \begin{subfigure}[t]{.24\linewidth}
            \includegraphics[width=\textwidth]{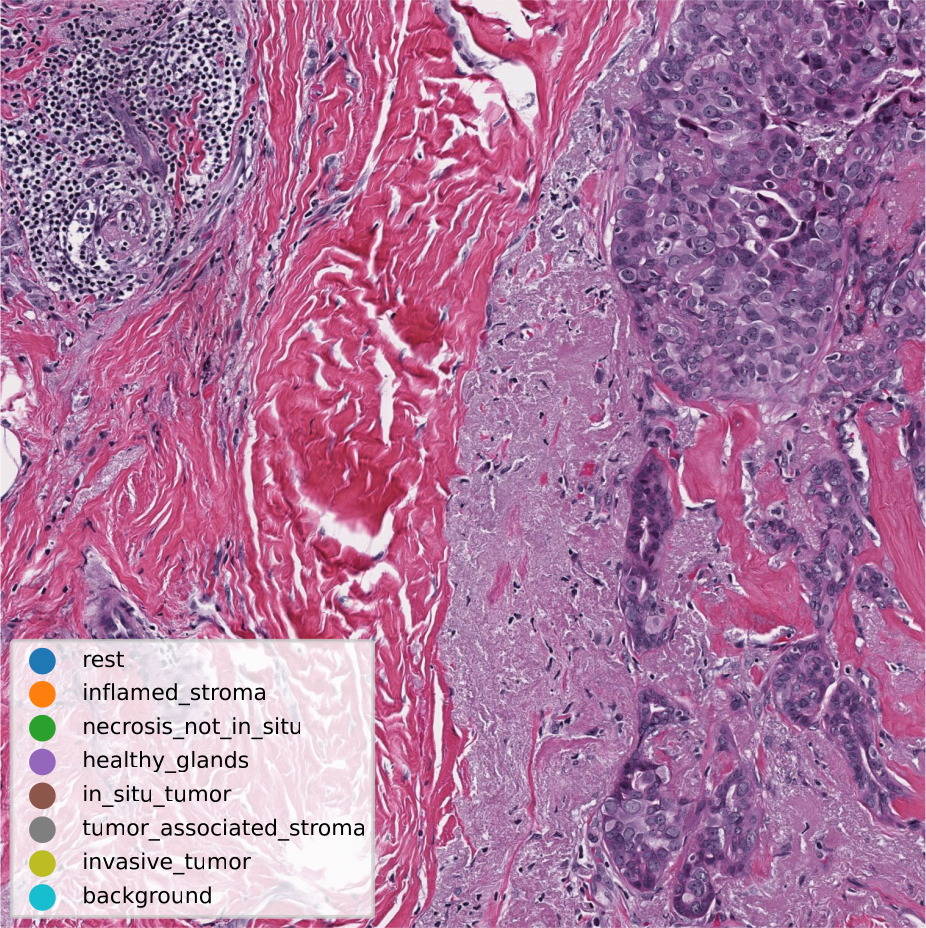}
        \end{subfigure}\hfill
        \begin{subfigure}[t]{.24\textwidth} 
            \includegraphics[width=\textwidth]{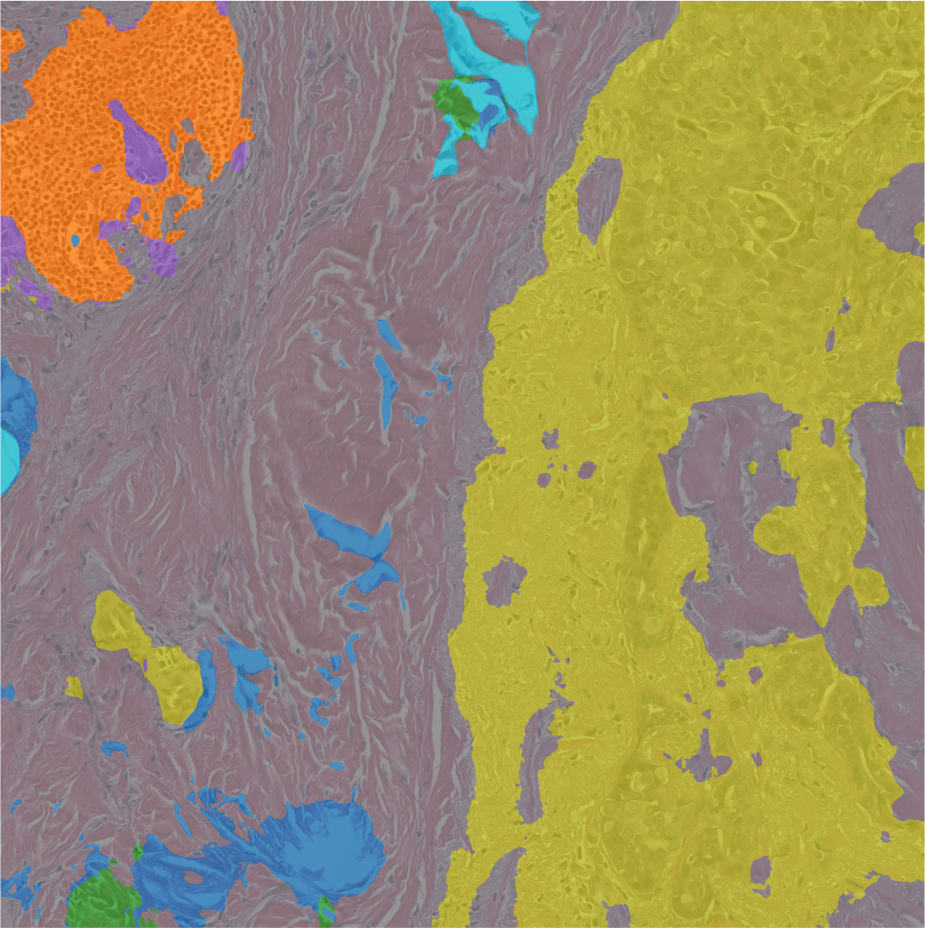}
            \caption{\begin{tabular}{l}
                 pooling (3)  \\
                 $0.5876\pm0.3430$
            \end{tabular}}
        \end{subfigure}\hfill
        \begin{subfigure}[t]{.24\textwidth}
            \includegraphics[width=\textwidth]{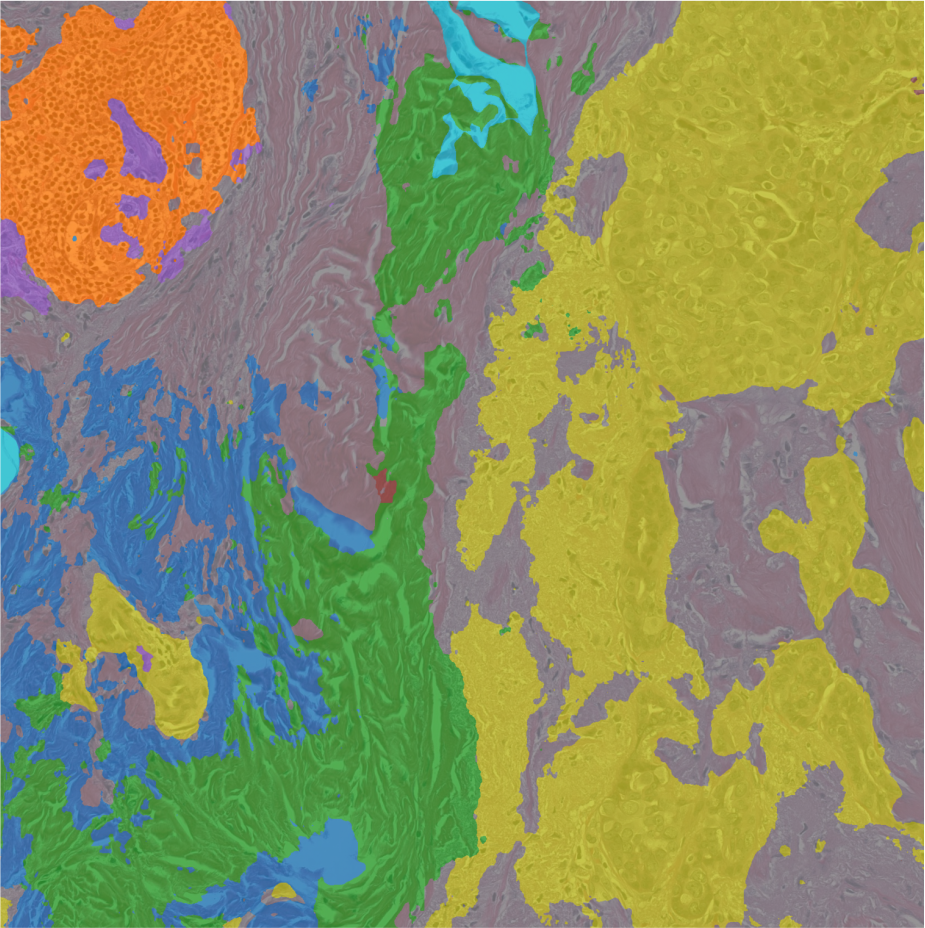}
            \caption{\begin{tabular}{l}
                 grouped conv (7)  \\
                 $0.5569\pm0.3372$
            \end{tabular}}
        \end{subfigure}\hfill
        \begin{subfigure}[t]{.24\textwidth}
            \includegraphics[width=\textwidth]{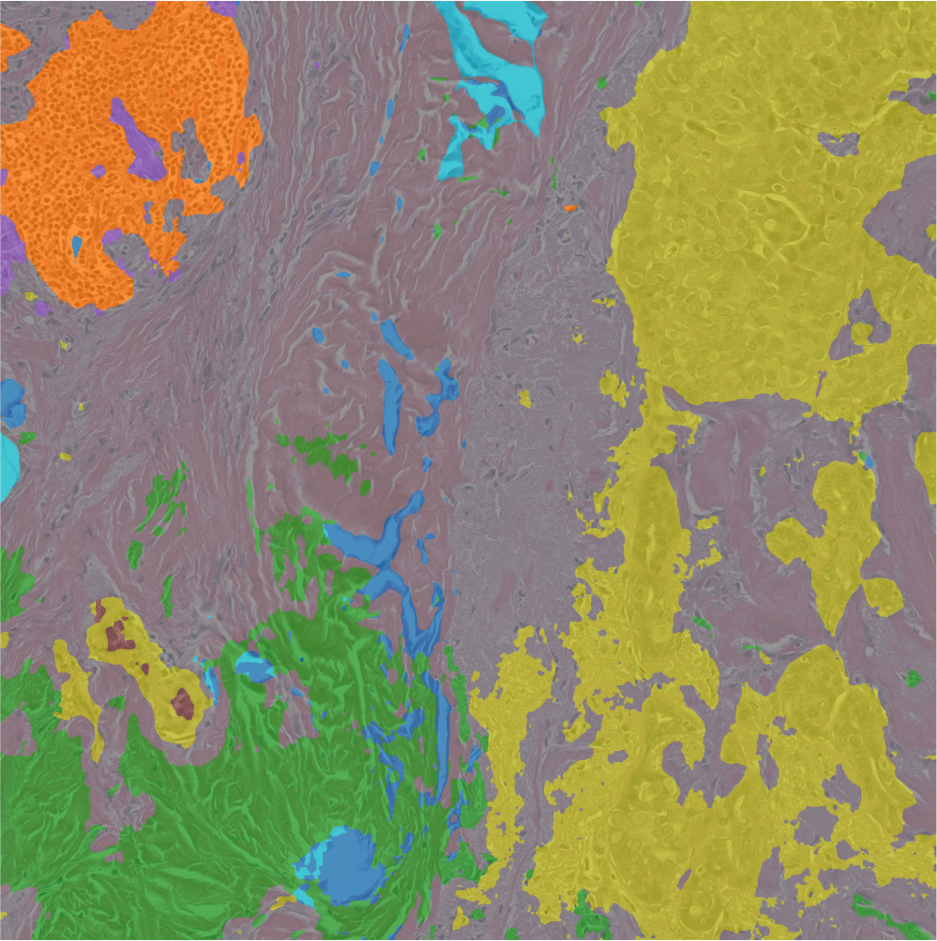}
            \caption{\begin{tabular}{l}
                 random  \\
                 $0.6142\pm0.3570$
            \end{tabular}}
        \end{subfigure}
    \end{subfigure}
    \caption{Qualitative segmentation results for a selection of token mixers (kernel size) on the 2D datasets. The shown sample is the median case of the mean \ac{dsc} over all token mixers. The captions state the \ac{dsc} ($\mu\pm\sigma$) of the case. Segmentation ground truth for the TIGER test case is shown separately shown in Fig.~\ref{fig:tiger_seg_ground_truth}.}
    \label{fig_ap:qualitative_segmentation_results}
\end{figure*}
\end{document}